\documentclass[ijoc,nonblindrev]{informs3} 
\usepackage{tikz}
\usetikzlibrary{positioning,shapes.geometric}
\usepackage{enumerate}
\usepackage{longtable}
\usepackage{bm}
\usepackage{vcell}
\usepackage{graphicx}
\usepackage{adjustbox}
\usepackage{algorithm}
\usepackage{algpseudocode}

\OneAndAHalfSpacedXI 
\usepackage{booktabs}
\usepackage{multirow}
\usepackage{arydshln}
\usepackage{rotating}
\usepackage{hyperref}
\usepackage{subfigure}
\usepackage{pgfplots}
\usepackage{comment}
\usepackage{dsfont}

\newtheorem{property}{Property}
\newtheorem{result}{Result}

\pgfplotsset{compat=newest}
\usepackage{caption}
\newlength{\commentindent}
\usepackage{natbib}
 \bibpunct[, ]{(}{)}{,}{a}{}{,}%
 \def\bibfont{\small}%
\TheoremsNumberedThrough  

\EquationsNumberedThrough 

\begin{document}

\RUNAUTHOR{Zhao et al.}

\RUNTITLE{Large Neighborhood and Hybrid Genetic Search for Inventory Routing Problems}

\TITLE{Large Neighborhood and Hybrid Genetic Search for Inventory Routing Problems}

\ARTICLEAUTHORS{%
\AUTHOR{Jingyi Zhao*\footnotetext{*Corresponding author}}
\AFF{
{Shenzhen Research Institute of Big Data, Shenzhen 518172, China}, \EMAIL{jingyi.z@u.nus.edu}}
 \AUTHOR{Claudia Archetti}
\AFF{{Department of Economics and Management,
 University of Brescia,  Italy}, \EMAIL{claudia.archetti@unibs.it}}
\AUTHOR{Tuan Anh Pham}
\AFF{{Faculty of DS\&AI, College of Technology, National Economics University, Vietnam}, \EMAIL{ptanh@neu.edu.vn}} 
\AUTHOR{Thibaut Vidal}
\AFF{{CIRRELT \& SCALE-AI Chair in Data-Driven Supply Chains, Department of Mathematical and Industrial Engineering,
 Polytechnique Montreal, Canada}, \EMAIL{thibaut.vidal@polymtl.ca}}
} 

\ABSTRACT{%
The inventory routing problem (IRP) focuses on jointly optimizing inventory and distribution operations from a supplier to retailers over multiple days. Compared to other problems from the vehicle routing family, the interrelations between inventory and routing decisions render IRP optimization more challenging and call for advanced solution techniques. A few studies have focused on developing large neighborhood search approaches for this class of problems, but this remains a research area with vast possibilities due to the challenges related to the integration of inventory and routing decisions. In this study, we advance this research area by developing a new large neighborhood search operator tailored for the IRP. Specifically, the operator optimally removes and reinserts all visits to a specific retailer while minimizing routing and inventory costs. We propose an efficient tailored dynamic programming algorithm that exploits preprocessing and acceleration strategies. The operator is used to build an effective local search routine, and included in a state-of-the-art routing algorithm, i.e., Hybrid Genetic Search. Through extensive computational experiments, we demonstrate that the resulting heuristic algorithm leads to solutions of unmatched quality up to this date, especially on large-scale benchmark instances.
}%



\maketitle

\section{Introduction} 
\label{intro}

The use of large neighborhood search (LNS) techniques has permitted significant progress in the solution of various vehicle routing problems (VRPs) \citep[see, e.g.,][]{christiaens2020slack,pacheco2023exponential} and combinatorial optimization (CO) problems in general \citep{hendel2022adaptive}. Especially when multiple decisions are interrelated, classical local search neighborhoods often do not allow an efficient search space exploration, while larger neighborhoods can be instrumental in exploring structurally different solutions.

The Inventory Routing Problem (IRP), in particular, is a prototypical example of a problem involving multiple interrelated decisions. In its canonical form, the IRP requires jointly optimizing inventory and delivery routes from a supplier to geographically dispersed retailers over a planning horizon measured in days.
This involves a complex set of decisions that include the daily selection of retailers receiving visits, their delivery quantities, the assignment of vehicles to these visits, and their ordering. Despite a few recent and promising works on the topic \citep[see, e.g.,][]{chitsaz2019unified,diniz2020efficient,archetti2021kernel,skaalnes2023branch}, applications of LNS for this problem remain few and far. Many LNS designed for IRP fundamentally rely on mixed integer programming \citep[see, e.g.,][]{bouvier2024solving}, which can become time-consuming and not permit many iterations.
For this reason, though extensive progress has been made on solution methods over the past decades, judging from the average gap and standard deviation of state-of-the-art methods, significant improvements are still possible on medium and larger-scale instances such as the ones introduced in \citet{skaalnes2024new}.

In this study, we contribute to advancing efficient large-neighborhood-based search techniques for the IRP. Our contribution is fourfold.
\begin{enumerate}
\item We propose a new large neighborhood operator that removes and reinserts all visits to a specific retailer while minimizing routing and inventory costs. While this may evoke the concepts used in ruin-and-recreate methods \citep{ropke2006adaptive}, the way we apply this strategy is fundamentally different. 
We demonstrate that, after preprocessing, the reinsertion can be done optimally
using efficient dynamic programming and acceleration strategies specifically tailored to this case.
Moreover, the speed of our method permits us to explore this large neighborhood fully for all retailers in random order until a local minimum is reached.
\item We integrate this LNS operator into the hybrid genetic search (HGS) of \citep{vidal2012hybrid,vidal2022hybrid}, extending this method to solve IRPs. Specifically, the new operator replaces the neighborhood search operator originally used in \citet{vidal2012hybrid} to choose the allowed visit-day combinations by inspection in the context of multi-period VRP.
This is the first time an operator has been designed for an IRP that optimizes replenishment decisions across the entire planning horizon.
\item We extend the IRP to the Stock-Out Allowed (SOA) model by introducing a penalty term for stock-out.
This variant has been rarely considered, but it is especially relevant when applying IRP solution techniques in a dynamic environment, e.g., as studied in \citep{greif2024combinatorial}, as this often requires the ability to evaluate infeasible solutions in some scenarios.
\item We conduct an extensive computational campaign to evaluate the neighborhood search efficiency and compare the adapted HGS's performance with other state-of-the-art IRP algorithms. Most notably, the resulting method identifies 187 new best solutions out of 800 small data instances and 178 new best solutions in 240 large instances. Even more remarkable is its performance when applied to the new challenging IRP dataset proposed in \cite{skaalnes2024new}, where it finds 191 new benchmark solutions on 270 instances with a shorter average running time. These findings highlight the efficiency of the proposed enhanced HGS algorithm, thanks to the development of the new large neighborhood operator.
 \end{enumerate}

\section{Related works} 
\label{sectionre}

\noindent
\textbf{Large Neighborhood Search (LNS)} is a fundamental component of heuristic algorithms and is regularly used in combinatorial optimization (CO), especially on large-scale problems of a size that defies exact algorithms. Broadly defined, a neighborhood search involves an iterative exploration of the solution space around an incumbent solution, i.e., a ``neighborhood", to find improved solutions. Typically, the search starts from an initial solution, and it iterates by replacing the incumbent solution with any superior solution found (i.e., accepting a ``move''), with mechanisms in place to allow controlled deterioration of the solution quality to escape local optima. The qualifiers ``local'' and ``large'' distinguish the scale of the neighborhood examined during each iteration. When the number of possible moves is polynomial, enabling efficient enumeration, the process is generally termed Local Search (LS). Conversely, when each iteration involves the exploration of a neighborhood with an exponential number of solutions, we typically refer to the process as an LNS. The neighborhood exploration methods might be different; they may be based on sampling or on efficient (possibly polynomial-time) search techniques tailored to the problem, often relying on combinatorial subproblems solvable in polynomial time.

\citet{ahuja2002survey} reviews various large neighborhood search techniques, categorizing them into three main categories. The first category, variable-depth methods, employs heuristics to explore increasingly large (higher-order) neighborhoods. The heuristic of \cite{lin1973effective} is a seminal example, as it effectively harnesses pruning techniques to explore many solutions obtained by replacing a variable number of edges. Applications of this technique to various CO problems are demonstrated in \citet{applegate2003traveling} and \citet{hendel2022adaptive}.
The second category includes network flow-based improvement algorithms, which rely on solving network flow subproblems to explore large neighborhoods. Ejection chains \citep{Thompson1993,Vidal2017b} and assignment-based neighborhoods \citep{Toth2008,Capua2018} are notable examples, leveraging shortest path and assignment problems to explore an exponential subset of solutions in polynomial time. The third category, neighborhoods induced by restrictions, exploits special cases of the original problem that admit polynomial-time solutions. For example, certain restrictions of the Traveling Salesman Problem (TSP) can be solved in polynomial time, leading to various classes of large neighborhoods that are broadly applicable to routing problems \citep{Balas2001,pacheco2023exponential}. Additionally, complex CO problems can often be decomposed into several classes of decision variables, allowing for local searches on a subset of these variables. When combined with a move evaluation procedure that optimally determines the remaining variables, this approach can lead to effective searches in large neighborhoods \citep{Vidal2017b,Toffolo2019}. Collectively, these techniques have led to important breakthroughs for difficult problems.\\

\noindent
\textbf{The Inventory Routing Problem (IRP)} is a foundational problem in supply chain optimization that has been extensively studied over the past decades, \textcolor{black}{as highlighted in surveys by \citet{anderssoN2010industrial}, \citet{coelho2014survey} and \citet{vidal2020survey}}. The IRP is a canonical problem designed to capture the challenges associated with the simultaneous optimization of both inventory replenishment strategies and routing plans, offering significant potential for cost reduction in supply chain operations.

The origins of the IRP trace back to the work of \citet{bell1983improving}, who introduced the concept for a gas delivery application. Their approach leveraged Lagrangian relaxation to decompose the problem across temporal and vehicle dimensions. Later, \citet{christianseN1999decomposition} expanded the IRP’s application to maritime logistics, coining the term ``inventory pickup and delivery problem.'' Their approach used Dantzig-Wolfe decomposition and column-generation techniques.  Further advancements in addressing the IRP were made by \citet{carter1996solving} and \citet{campbell2004decomposition}, who developed efficient heuristic methods by decomposing the IRP into an allocation problem and a routing problem.

Numerous metaheuristic strategies have since been proposed for the IRP. Examples include tabu search \citep{rusdiansyah2005integrated}, genetic algorithms \citep{abdelmaguid2006genetic}, a greedy randomized adaptive search procedure \citep{campbell2004decomposition}, adaptive large neighborhood search (LNS) \citep{coelho2012inventory}, and a variable neighborhood descent heuristic with stochastic neighborhood ordering \citep{alvarez2018iterated}, which combines routing-focused and joint routing-inventory operators. In addition, ``matheuristics'', which integrate heuristic search with mathematical programming, have shown significant promise. Notable examples include hybrid methods by \citet{archetti2012hybrid,archetti2017matheuristic,chitsaz2019unified,archetti2021kernel,skaalnes2023branch}, \citet{Bouvier2024} and \citet{dinh2025matheuristic}.
Among these, \citet{skaalnes2023branch} developed a matheuristic that is invoked on each primal solution identified by a branch-and-cut algorithm. Whenever the mathematical program discovers a new best upper bound, this method attempts to further improve it by applying mixed integer programming to a restricted problem, which involves removing and reinserting clusters of clients. This advanced approach successfully identified 741 best-known solutions out of 878 existing benchmark instances and proved the optimality of 458 of these solutions. 

Research has also extended to various IRP variants. For instance, \citet{gaur2004periodic} addressed the periodic IRP, optimizing replenishment under cyclical demand, while other studies focused on cyclic \citep{gunawaN2019simulated, vincent2021multi}, multi-product \citep{ohmori2021multi, mahjoob2022modified},  multi-depot \citep{guimaraes2019two}, \textcolor{black}{multi-attribute \citep{coelho2020variable}, and bi-objective \citep{feng2025bi} IRPs.} Strategic extensions integrating production decisions have also been explored \citep{adulyasak2014optimization} As noted by \citet{anderssoN2010industrial}, early research predominantly relied on heuristics due to the problem’s complexity. However, subsequent works introduced exact algorithms, primarily branch-and-cut methods \citep{CoelhoLaporte2013,CoelhoLaporte2013b,adulyasak2014formulations,Archetti2014,CoelhoLaporte2014}, with one notable exception employing branch-and-price \citep{DesaulniersRC2016}. \textcolor{black}{Other contributions include VND methods \citep{larrain2017variable} and matheuristic frameworks \citep{vincent2025three} that combine mathematical programming with metaheuristics to address larger-scale instances.}

Despite their efficiency for small to medium-scale problems, these methods often struggle with scalability, particularly for instances involving more than 50 retailers or planning horizons exceeding three days. Addressing this gap, \citet{skaalnes2024new} proposed challenging large-scale IRP instances to stimulate further research. Recently, \citet{Bouvier2024} introduced a large neighborhood search procedure designed for very large IRPs involving an average of 30 commodities, 16 depots, and 600 retailers. However, the scale of the problems they address and the complexity of the underlying neighborhoods limit the procedure to only a handful of local search descents.  This highlights the ongoing need for efficient neighborhood operators ---both local and large--- that can effectively tackle large-scale IRP instances.

\section{Problem Statement}
\label{seproblemstatement}

Our IRP is defined on a complete directed graph $\mathcal{G} = (\mathcal{N}, \mathcal{A})$, where the node set $\mathcal{N} = \{0, n+1\} \cup \mathcal{N}'$ includes the departure depot at node $0$, the destination depot at node $n+1$, and the retailer nodes $\mathcal{N}' = \{1, \ldots, n\}$. The arc set $\mathcal{A}$ represents paths between nodes, with each arc $(i, j) \in \mathcal{A}$ associated with a non-negative cost $c_{i,j}$. The planning horizon spans $H$ days, denoted by $\mathcal{T} = \{1, \ldots ,H\}$. Each retailer node $i \in \mathcal{N}'$ has a per-day inventory holding cost $h_i$, an initial inventory level $I_i^0$, and a maximum inventory level $U_i$. In the Stock-Out Allowed (SOA) model, $\rho h_i$ represents the per-unit stock-out penalty for retailer $i$, with $\rho > 1$. In contrast, in the No Stock-Out (NSO) model, the value of $\rho$ is set to infinity.
Each day $t \in \mathcal{T}$, $d_{0}^t$ units of product are made available at the supplier, and $d_i^t$ units are consumed by retailer $i \in \mathcal{N}'$. A fleet $\mathcal{K}$ of $K$ identical vehicles, each with capacity $Q$, provides the delivery service.

The IRP involves creating a distribution plan that minimizes the total cost. This plan specifies, for each day $t \in \mathcal{T}$, the delivery quantity to each retailer $i \in \mathcal{N}'$ and the routes of vehicles for servicing the retailers on that day. The total cost encompasses the holding costs at both the supplier and retailers, the stock-out costs under the SOA model, and the transportation costs associated with the routes.
\textcolor{black}{In the IRP, allowing stock-outs is a realistic modeling approach that reflects many real-world scenarios. This no-backlogging setup is particularly suited to situations where unmet demand cannot or should not be fulfilled later. For example, in fast-moving consumer goods, customers typically turn to competitors if products are out of stock, meaning lost sales rather than deferred ones. Similarly, in retail, pharmaceuticals, or seasonal product delivery, stock-outs represent immediate service failure, and backlogging is irrelevant or impractical. The penalty cost serves as a proxy for lost revenue, customer dissatisfaction, or contractual penalties, and it incentivizes timely and adequate deliveries. This modeling choice also simplifies the mathematical formulation by avoiding the complexity of backlog tracking while maintaining economic pressure to avoid stock-outs.}

A mathematical formulation of the IRP under the SOA model is presented below, extending the formulation from \citet{archetti2007branch} to account for possible stock-outs. Let $I_i^t$ represent the inventory level at retailer $i$ at the end of the day $t$, and $B_i^t$ the unsatisfied demand on that day. Also, $I_0^t$ is the  inventory level at the supplier at the end of the day $t$. Binary variable $z^{k,t}_{i}$ is set to one if node $i \in \mathcal{N}$ is visited by vehicle $k \in \mathcal{K}$ on day $t \in \mathcal{T}$. Continuous variable $q^{k,t}_{i}$ denotes the quantity delivered to retailer $i \in \mathcal{N'}$ on day $t$ by vehicle $k$. Binary variable $y^{k,t}_{i,j}$ indicates whether vehicle $k$ traverses arc $(i, j) \in \mathcal{A}$ on day $t$. Finally, $Q_i^{k,t}$ is a  continuous variable representing the load of vehicle $k$ upon arrival at node $i$, reflecting the cumulative load at that point in the vehicle’s route.
The objective is to minimize the total cost, which includes holding costs at the supplier $h_0I_{0}^t$, holding costs at the retailers $\sum_{i \in \mathcal{N'}}h_iI_i^t$, penalty costs for stock-outs $\sum_{i \in \mathcal{N'}} \rho h_i B_i^t$, and transportation costs $\sum_{k \in \mathcal{K}} \sum_{(i,j) \in \mathcal{A}} c_{i,j}y^{k,t}_{i,j}$. The resulting mathematical formulation~is:
\begin{equation}
\min \sum_{t \in \mathcal{T}} \bigg( h_0I_{0}^t + \sum_{i \in \mathcal{N'}}h_iI_i^t +\sum_{i \in \mathcal{N'}} \rho h_i B_i^t + \sum_{k \in \mathcal{K}} \sum_{(i,j) \in \mathcal{A}} c_{i,j}y^{k,t}_{i,j} \bigg)\label{eq:1a} 
\end{equation}
\begin{align}
 \text{s.t.} \quad & I_0^t = I_{0}^{t-1} + d_{0}^t - \sum_{k \in \mathcal{K}} \sum_{i \in \mathcal{N'}} q^{k,t}_i\quad & \forall\ t \in \mathcal{T,} \label{eq:1b} \\
 & I_i^t = I_i^{t-1} - d_i^t + \sum_{k \in \mathcal{K}} q^{k,t}_i +B_i^t \quad & \forall\ i \in \mathcal{N'},\ t \in \mathcal{T,} \label{eq:1c} \\
 & \sum_{k \in \mathcal{K}} q^{k,t}_i + I_i^{t-1} \leq U_i \quad & \forall\ i \in \mathcal{N'}, t \in \mathcal{T}, \label{eq:1e} \\
 & q^{k,t}_i \leq U_iz^{k,t}_i \quad & \forall\ i \in \mathcal{N'}, k \in \mathcal{K}, t \in \mathcal{T}, \label{eq:1f} \\
 & \sum_{i \in \mathcal{N'}} q^{k,t}_i \leq Qz^{k,t}_{0} \quad & \forall\ k \in \mathcal{K}, t \in \mathcal{T}, \label{eq:1g} \\
 & \sum_{k \in \mathcal{K}} z^{k,t}_i \leq 1 \quad & \forall\ i \in \mathcal{N'}, t \in \mathcal{T}, \label{eq:1h} \\
 & \sum_{j \in \mathcal{N'}\cup \{n+1\} } y^{k,t}_{{i,j}} = \sum_{j \in \mathcal{N'}\cup \{0\}} y^{k,t}_{{ji}} = z^{k,t}_i \quad & \forall\ i \in \mathcal{N'}, k \in \mathcal{K}, t \in \mathcal{T}, \label{eq:1i} \\
 & \sum_{j \in \mathcal{N'} \cup \{n+1\} } y^{k,t}_{{0,j}} =z_0^{k,t}= \sum_{j \in \mathcal{N'}\cup \{0\}} y^{k,t}_{{j,n+1}}=z_{n+1}^{k,t} =1 \quad & \forall\ k \in \mathcal{K}, t \in \mathcal{T}, \label{eq:same} \\
 & Q_0^{k,t}= \sum_{i \in \mathcal{N'}} q_i^{k,t} \quad & k \in \mathcal{K}, t \in \mathcal{T}, \label{eq:load0}\\
 &Q_i^{k,t}-q_i^{k,t}\geq Q_j^{k,t}-Q \bigg(1-y^{k,t}_{i,j} \bigg) \quad & \forall\ (i, j) \in \mathcal{A}, k \in \mathcal{K}, t \in \mathcal{T}, \label{eq:subtour} \\
 & Q^{k,t}_i,q^{k,t}_i \geq 0 \quad & \forall\ i \in \mathcal{N'}, k \in \mathcal{K}, t \in \mathcal{T}, \label{eq:1l} \\
 & q^{k,t}_0 = q^{k,t}_{n+1} = 0 \quad &\forall\ k \in \mathcal{K}, t \in \mathcal{T}, \label{eq:q0} \\
 & z^{k,t}_i \in \{0, 1\} \quad & \forall\ i \in \mathcal{N'}, k \in \mathcal{K}, t \in \mathcal{T}, \label{eq:1k} \\
 & I ^{t}_i \geq 0, B^{t}_i \geq 0 \quad & \forall\ i \in \mathcal{N'}, t \in \mathcal{T}, \label{eq:hb} \\
 & y^{k,t}_{i,j} \in \{0, 1\} \quad & \forall\ (i, j) \in \mathcal{A}, k \in \mathcal{K}, t \in \mathcal{T}.\label{eq:1m} 
    \end{align}

Constraints~\eqref{eq:1b} define the inventory level at the supplier
while constraints~\eqref{eq:1c} determine the inventory level at retailer $i$ at the end of the day $t$, where $B_i^t$ represents the stock-out quantity. 
Constraints~\eqref{eq:1e} ensure that the total quantity delivered to retailer $i$ on day $t$ by all vehicles $k \in \mathcal{K}$, combined with the previous day's inventory $I_i^{t-1}$, does not exceed the retailer's inventory capacity $U_i$. This follows the maximum level policy, which allows flexibility in delivery quantities as long as post-delivery levels remain within maximum inventory limits. Constraints~\eqref{eq:1f} ensure that the delivered quantity is zero if the vehicle does not visit the retailer ($z^{k,t}_i = 0$). Constraints~\eqref{eq:1g} impose vehicle capacity limits, while constraints~\eqref{eq:1h}--\eqref{eq:q0} model the routing constraints. Specifically, constraints~\eqref{eq:1h} stipulate that each retailer can be visited no more than once per day. Constraints~\eqref{eq:1i} regulate the flow through each node for each vehicle on each day. Constraints~\eqref{eq:same} require each vehicle to depart from and return to the depot, while constraints~\eqref{eq:load0}--\eqref{eq:q0} ensure proper tracking of the vehicle load throughout its journey and prevent subtours.

Finally, Constraints~\eqref{eq:hb}, in conjunction with the objective function, ensure that $I_i^t$ is set to zero when inventory is insufficient to meet demand and that $B_i^t$ remains zero when there is no stock-out.

The following property holds and will play a role in the analysis of the complexity of the large neighborhood operators we introduce:
\begin{property}[\cite{archetti2022comparison}]\label{obs:integer}
     When the input parameters $Q$, $U_i$, and $d_i^t$ are integer, there always exists an optimal solution in which the quantity variables $q_i^{k,t}$ (and, consequently, the inventory and stock-out variables $I_i^t$ and $B_i^t$) are integer-valued for all $i \in \mathcal{N}$, $k \in \mathcal{K}$, and $t \in \mathcal{T}$.
\end{property}

\section{A New Large Neighborhood Search Operator}
\label{edu:DSopt}

This paper proposes a new, efficient neighborhood search operator for the Inventory Routing Problem (IRP), designed to be broadly applicable, regardless of the metaheuristic context. Specifically, given an incumbent solution and a selected retailer~$i$, it removes all visits to it over the planning horizon and optimally selects new visit days, delivery quantities, and delivery locations among existing routes or new direct routes with a single visit, while accounting for all relevant costs.  

Consequently, this move \textcolor{black}{can also be viewed as a simultaneous reinsertion operator} across multiple days, specifically designed for IRP. It is iteratively applied to all retailers in random order until a local minimum is reached.
The closest comparable concept in the literature is the \textit{pattern improvement} (PI) operator proposed in \cite{vidal2012hybrid}, which optimally selects a visit pattern over an extended planning horizon and insertion positions for the multi-period VRP. However, the operator introduced here for the IRP is significantly different, as it handles inventory level constraints, accounts for possible stock-outs, and solves an underlying combinatorial optimization problem to identify the best reinsertion configuration. As a result, it will be referred to as the \textit{delivery schedule} (DS) operator throughout the remainder of this document. As will be shown, dynamic programming (DP) algorithms can be designed to find optimal reinsertion positions and quantities. The proposed method is versatile and can be adapted to different IRP variants, such as those with penalized stock-outs.

We note that the DS operator allows the insertion of a visit to a retailer into a route, even if this causes a violation of the capacity constraints. The capacity violation is penalized by multiplying the excess quantity by a penalty parameter $\omega$. The insertion procedure, together with the definition of the quantity to be delivered and the corresponding cost, is detailed in Section~\ref{preprocess}.

 \begin{figure}[htbp]
\hspace*{-0.5cm}
\includegraphics[width=1.05\textwidth]{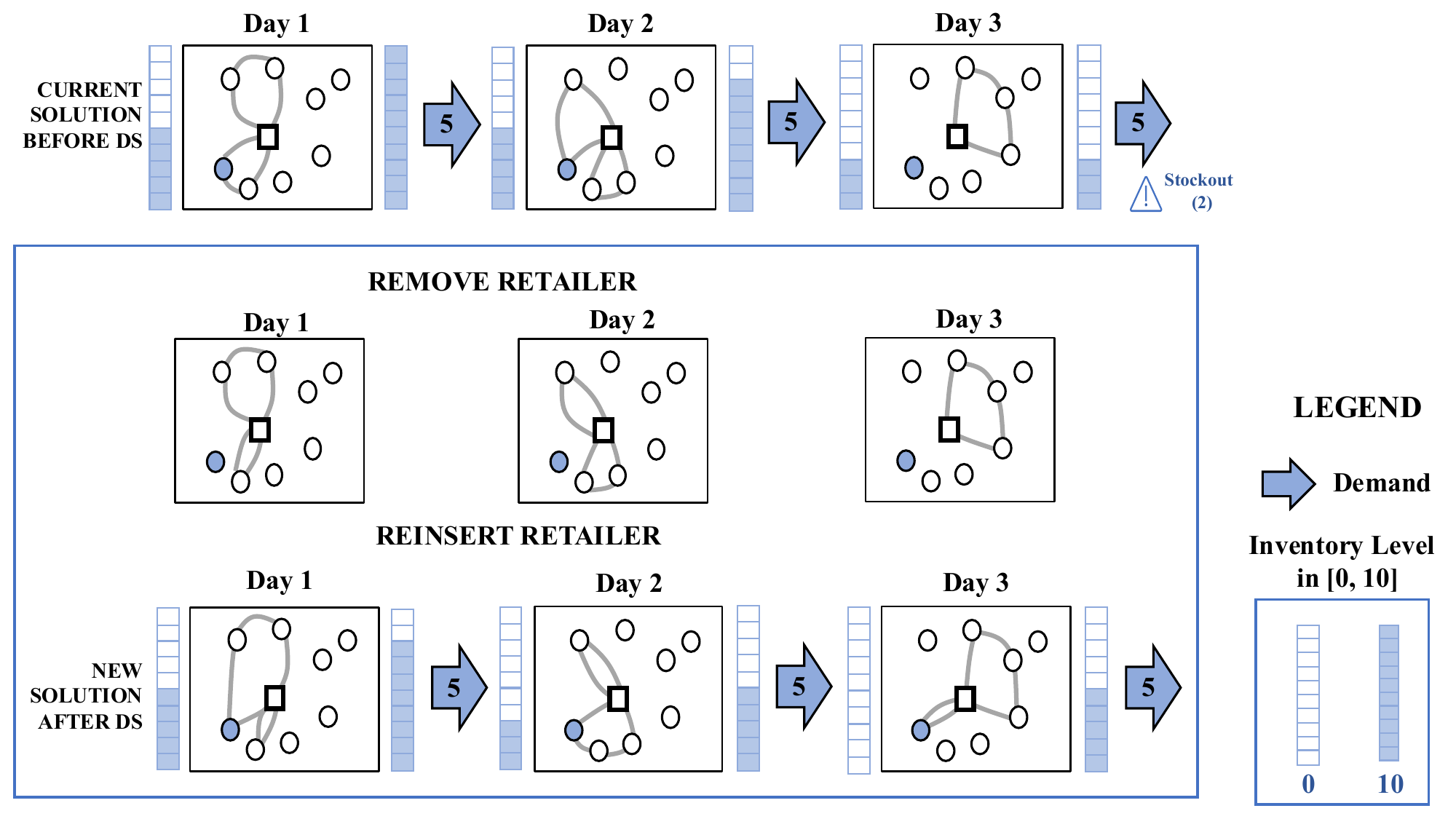}
\caption{Illustration of the DS neighborhood} 
\label{fig:in}
\end{figure}

Figure~\ref{fig:in} illustrates how the DS operator works. One of the retailers (highlighted in blue in the figure) is selected to be removed from its original visit positions, as shown in the first row. 
After applying the DP algorithm explained below, new insertion positions are determined, as depicted in the second row. Note that the delivery quantities for each day have also been adjusted, leading to completely different inventory levels over the planning horizon. Importantly, while the initial situation is associated with a stock-out quantity equal to $2$ in the third period, the move results in a solution with no stock-outs, as illustrated in the bars close to each plot.

\subsection{Dynamic Programming Algorithm}
\label{DProutines}

After having removed a retailer $i$ from all routes, the DS operator determines the best reinsertions in terms of days, routes, and quantities. The evaluation of the reinsertion cost in the IRP is quite challenging. Indeed, the cost depends on the inventory level, which in turn depends on the quantity delivered in former visits to the retailer. Our main contribution is the design of an efficient method to compute the reinsertion cost function. To do that, we first observe that, for each day $t$, the reinsertion cost function is a piecewise linear function in the inventory level $I_i^t$. Then, we develop a DP algorithm that iteratively computes an optimal replenishment strategy.

The DP algorithm efficiently computes function $C_t(I_i^t)$, which represents the minimum cost associated with inventory level $I_i^t$ for retailer $i$ at time $t$, from $t=1$ to $t=H$. The cost includes inventory cost or stock-out cost at the retailer, detour cost associated with  the insertion of retailer $i$ in the daily routes of day $t$ (in case a positive quantity is delivered at time $t$), supplier's holding costs (which is negative in case of positive quantity delivered), and penalties associated with vehicle capacity violations. Since the inventory level is represented at the end of the day, after deduction of the retailer demand quantity $d^t_i$ on that day, and the inventory at the start of the day is upper limited to $U_i$ due to capacity limits, we have $I_i^t \in [0,U_i-d^t_i]$.

\begin{figure}[htbp]
\hspace* {-1cm}
\centering
\begin{overpic}[width=1.1\textwidth]{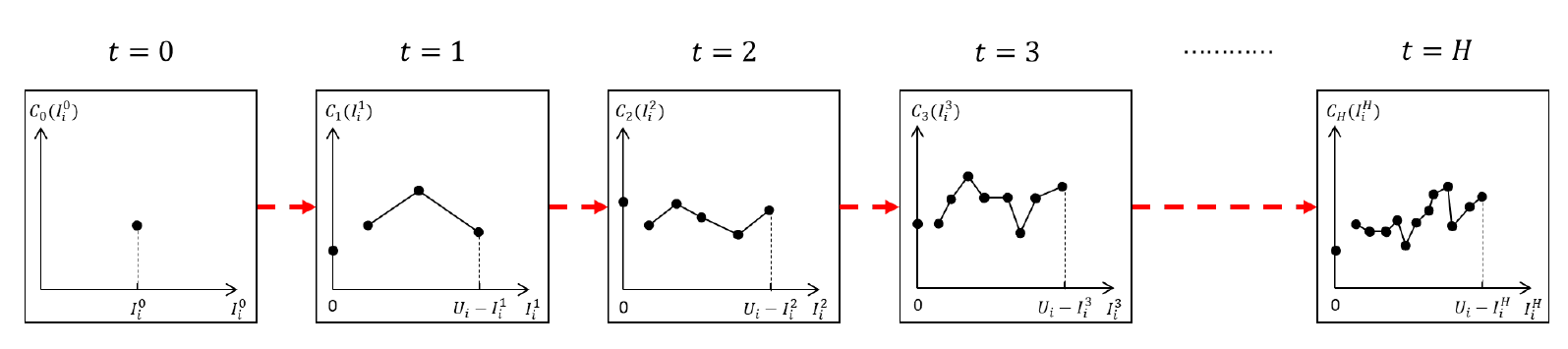}
\end{overpic}
\caption{The evolution of the cost function $C_t(I_i^t)$} 
\label{fig:DP_graph}
\end{figure}

Functions $C_t(I_i^t)$ are piecewise linear, and it is possible to compute them recursively using DP, from $t-1$ to $t$, as illustrated in Figure \ref{fig:DP_graph}. In practice, these functions will be efficiently represented as linked lists of function pieces, ordered by increasing inventory value, each of which is represented by the endpoints and slope. Moreover, due to Property~\ref{obs:integer}, any segment that does not span an integer inventory value can be discarded throughout the algorithm.
Thus, the complexity of the DP algorithm will depend on the number of pieces of this function, as analyzed in Section \ref{sec:complexity}. 

The cost function $C_t(I_i^t)$ is computed in two steps for each $t$. First of all, we compute a delivery cost function $\hat{C}_t(\hat I^t_i)$ considering inventory levels $\hat I^t_i \in [-d_i^t,U_i - d_i^t]$ that are possibly negative and trigger up to $d_i^t$ units of stock-out at the current period $t$. Then, we include the related stock-out penalties to derive $C_t(I_i^t)$ for $I_i^t \in [0,U_i - d_i^t]$. Function $\hat{C}_t(\hat I^t_i)$, can be computed by taking the minimal cost out of two options:
    \begin{enumerate}
        \item[(i)] If no delivery is performed at time $t$, the inventory level $I_i^{t-1}$ can be obtained by $ I_i^{t-1}= \hat I_i^{t}+d_i^{t}$ for a given value of $\hat I^t_i$, thus the cost is:
        \begin{equation}
            f_1 (\hat I^t_i) = C_{t-1}(\hat I^t_i+d_i^t) + h_i \hat I^t_i, \label{eq:metho1}
        \end{equation}
         where $h_i \hat I^t_i$ is the holding cost.
        \item[(ii)] If a delivery is performed, we should select the quantity $q_i^t > 0$ for which the cost associated with $\hat I^t_i$ is minimized. Note that, for a given value of $\hat I^t_i$, once we determine the quantity $q_i^t$, the inventory level $I_i^{t-1}$ corresponds to $ I_i^{t-1}= \hat I_i^{t}+d_i^{t}-q_i^{t}$. Thus, the cost function associated with this case is:
        \begin{equation}
            f_2 (\hat I_i^t) = \min_{q_i^t > 0} \bigg(C_{t-1}(\hat I_i^{t}+d_i^t-q_i^t) + h_i\hat I_i^t + F_t(q_i^t) - h_0^t(H-t+1)q_i^t  \bigg), \label{eq:metho2}
        \end{equation}
         where $F_t(q_i^t)$ is a piecewise linear function calculating the necessary detour (and possible capacity penalty) costs to insert a visit to retailer $i$ into a route on day $t$ and deliver $q_i^t$ units, and $h_0^t(H-t+1)q_i^t$ is the reduction in holding cost at the supplier for the remaining days. More details on the pre-processing of function $F_t(q_i^t)$ are given in the next section.
    \end{enumerate}
    Overall, the cost $\hat{C}_t(\hat I_i^t)$ is calculated as:
    \begin{equation}
        \hat{C}_t(\hat I_i^t)= \min\bm(f_1( \hat I_i^t), f_2(\hat I_i^t)\bm).  \label{eq:metho3}
    \end{equation}
Finally, we must derive function $C_t(I_i^t)$ considering stock-out penalties. Note that, thanks to the way stock-out costs are defined ($\rho h_i$ with $\rho >1$), it is always convenient to avoid any unnecessary stock-out leading to $I_i^t > 0$ in any given day. Otherwise, when a stock-out occurs (i.e., $\hat I_i^t \leq 0$ in Equation~\eqref{eq:metho3}), a penalty of $\rho h_i$ multiplied by the number of stock-out units ($-\hat I_i^t$) is incurred, and the additional holding cost term $h_i \hat I_i^t$ in Equation~\eqref{eq:metho3} should be removed accordingly. Therefore, we can calculate the function as: 
\begin{equation}
C_t(I_i^t) =
    \begin{cases}
    \min_{\hat I_i^t \in [-d_i^t,0]} \left( \hat{C}_{t} (\hat I_i^t) -
            \rho h_i \hat I_i^t - h_i \hat I_i^t  \right) & \text{if} \ I_i^t = 0 \\
    \hat{C}_t(\hat I_i^t) & \text{otherwise}.
    \end{cases}  \label{eq:metho4}
\end{equation}


Observe that, when stock-outs are forbidden, we can take $C_t(I_i^t) = \hat{C}_t(\hat I_i^t)$ considering only cases where $\hat I_i^t \geq 0$. \textcolor{black}{Moreover, backlogging (allowing negative inventory levels to be carried over and fulfilled in future periods) is not considered in the present paper, but the DS operator could accommodate backorders with minimal structural changes as explained in Appendix~\ref{appendix:backlog}.}

Once the DP algorithm has reached day $H$, it determines the inventory level $I_i^{H*}$ associated with the minimum cost $C_H(I_i^H)$. i.e., $I_i^{H*}=\arg\min_{I_i^H} C_H(I_i^H)$. Finally, a backtracking procedure, starting from~$I_i^{H*}$, determines the quantity $q_i^t$ to be delivered to the retailer on each day~$t$.



\subsection{Preprocessing Functions $F_t(q_i^t)$}
\label{preprocess}

Before executing the DP algorithm for a retailer $i$, we preprocess the functions $F_t(q_i^t)$ for each day $t$, representing the minimum cost of inserting a visit to retailer $i$ to deliver $q_i^t$ units. This involves evaluating all existing routes on day $t$ to calculate their residual capacity and detour cost $(\kappa, \gamma)$ associated with a least-cost insertion. If fewer than $K$ routes exist on day $t$, we also consider creating a new route for $i$, with a detour cost $\gamma = c_{0i} + c_{i0}$ and residual capacity $\kappa = Q$.

\begin{figure}[htbp]
\centering
\begin{overpic}[width=0.5\textwidth]{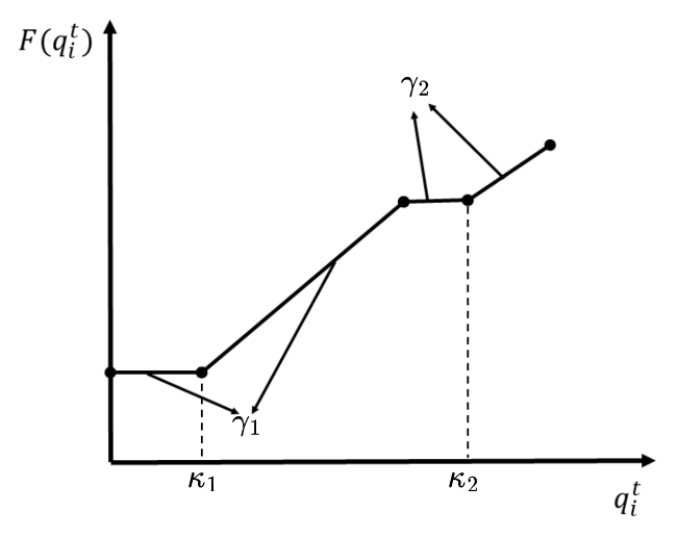}
\end{overpic}
\caption{Illustration of a function $F_t(q_i^t)$}
\label{fig:prep}
\end{figure}

In alignment with the penalty framework employed to diversify the search in \cite{vidal2012hybrid,vidal2022hybrid}, we allow penalized capacity violations in the routes, weighted by a penalty factor~$\omega$ \textcolor{black}{which remains fixed during each execution of the DS operator.} Based on these definitions, we have $F_t(q_i^t) = \min_{k\in\mathcal{K}} \left( \gamma_k + \omega \max(0, q_i^t - \kappa_k) \right)$. This function can be built in $O(K \log(K))$ time by ranking delivery options in increasing order of detour cost, resulting in a list $((\kappa_1, \gamma_1), \dots, (\kappa_m, \gamma_m))$ of $m \leq K$ delivery options, and eliminating dominated options. An illustration of the resulting function is presented in Figure~\ref{fig:prep}.

\subsection{Practical Implementation}
\label{acceleration}

\begin{figure}[htbp]
\centering
\includegraphics[width=\textwidth,]{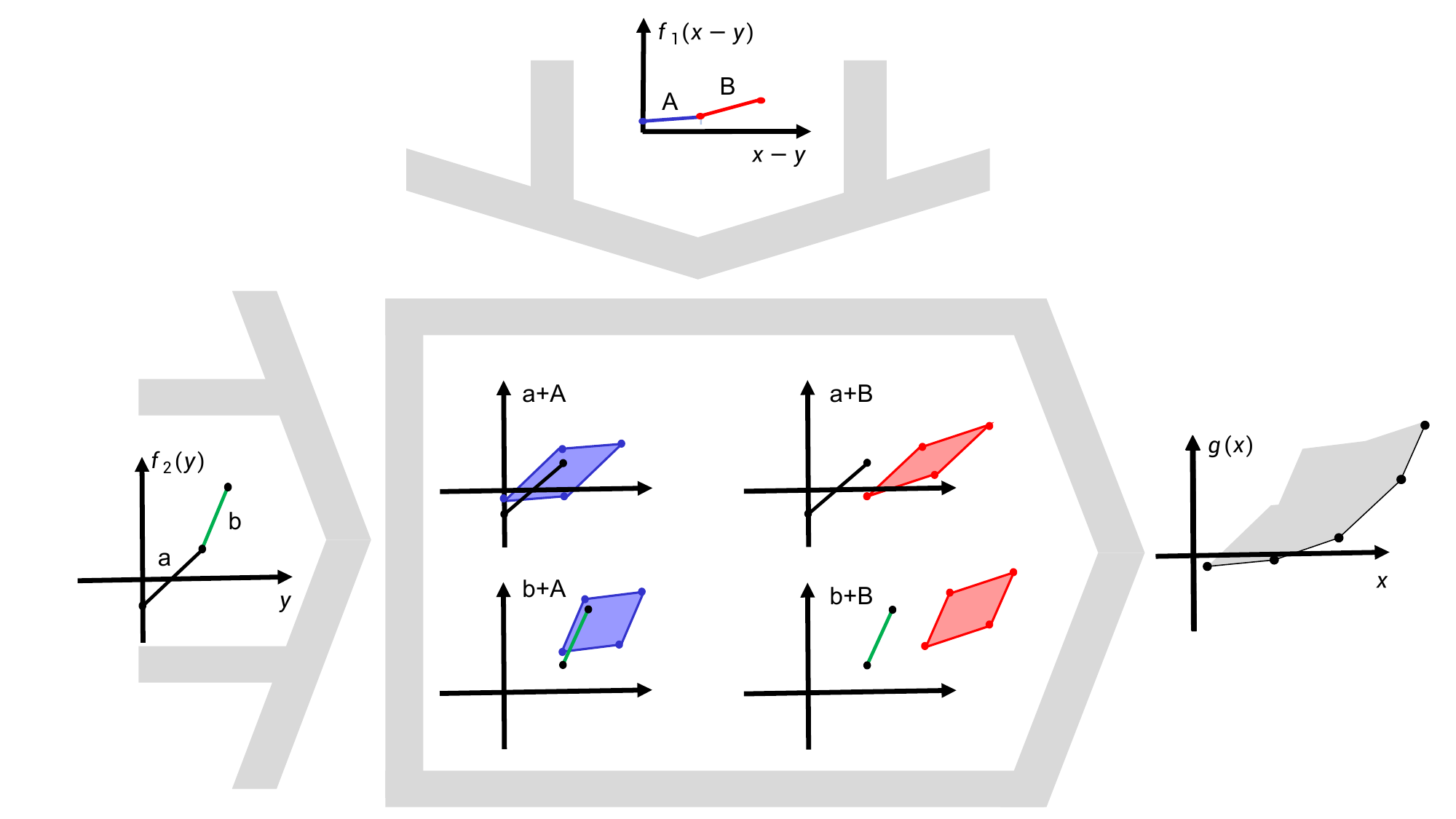}
\caption{An example of computation of the superposition of two piecewise linear functions. }
\label{fig:C34}
\end{figure}

As discussed previously, each function $C_t(I_i^t)$ and $F_t(q_i^t)$ is piecewise linear and can be efficiently represented as a linked list of function pieces represented by their interval and slope. A practical implementation of the calculations of Equations~(\ref{eq:metho1}--\ref{eq:metho4}) requires efficiently handling some primitive operations on these functions:
\begin{itemize}
    \item \textbf{Translation}: Compute $g(x) = f(x+a)$ where $a \in \Re$ is a constant. This operation occurs in Equations \eqref{eq:metho1} and \eqref{eq:metho2} and can be done in $O(\Phi(f))$ elementary operations by translating each segment, where $\Phi(f)$ is the number of pieces of $f$. It does not change the number of pieces of the resulting function.
    \item \textbf{Sum with an affine function}: Compute $g(x) = f(x) + a x$ with $a \in \Re$. This operation occurs in Equations \eqref{eq:metho1} and \eqref{eq:metho2}. It can be performed in $O(\Phi(f))$ elementary operations by changing the slope of each segment of $f$. The number of pieces of the function remains unchanged.
    \item \textbf{Lower envelope}: Compute $g(x) = \min(f_1(x), f_2(x))$. It occurs in Equation~\eqref{eq:metho3} and can be done in $O(\Phi(f_1) + \Phi(f_2))$ elementary operations by jointly sweeping through the segments of $f_1$ and $f_2$ and keeping track of the minimum, producing $O(\Phi(f_1) + \Phi(f_2))$ pieces.
    \item \textbf{Infimal convolution}: Compute $g(x) = \min_y(f_1(x-y) + f_2(y))$ \citep[see, e.g.,][]{Bauschke2017}. This operation is more complex and occurs during the calculations of Equation \eqref{eq:metho2}. It can be done in $O(\Phi(f_1) \Phi(f_2))$ elementary operations by observing that $g(x)$ is the lower envelope of the infimal convolutions of all pairs of segments in $f_1$ and $f_2$. Each such convolution can be geometrically calculated as the lower edges of a parallelogram, as illustrated in Figure~\ref{fig:C34}. The resulting function $g$ can have $O(\Phi(f_1) \times \Phi(f_2))$ pieces. 
\end{itemize}

\subsection{Computational Complexity}
\label{sec:complexity}

Preprocessing the functions $F_t(q_i^t)$ for all days requires $O(HK \log(K))$ time  (Section~\ref{preprocess}). Next, the computational effort to resolve the DP recursion depends on the number of pieces in the $C_t(I_i^t)$ functions. Since segments that do not contain integer values in the range $\{0, \dots, U_i\}$ can be disregarded, the number of pieces is in $O(U_i)$. 
The most computationally intensive operation in each recursion is the calculation of the infimal convolution between the functions $C_{t-1}$ and $F_{t}$, which have $O(U_i)$ and $O(K)$ pieces, respectively. This convolution requires $O(U_i K)$ elementary operations.  
However, many of these are later discarded, either because they fall outside the valid inventory range or due to domination when retaining the lower-cost segments.
As the convolution is performed for each of the $H$ days, exploring the DS neighborhood for a single retailer $i$, including preprocessing, takes $O(U_i K H)$ time. This analysis assumes $\log(K) \leq U_i$, which typically holds in practice. Moreover, as demonstrated in our computational experiments in Section~\ref{sec:complexity-experiment}, the number of pieces in the function $C_t$ remains far below their theoretical maximum in practice. For our problem instances, the complete evaluation of the DS neighborhood for any retailer $i$ is typically completed in a hundred microseconds.

\section{Hybrid Genetic Search}
\label{seml}

\textcolor{black}{The DS operator described above is integrated into the HGS framework initially introduced by \citet{vidal2012hybrid}, which we adapt to solve the IRP. While the DS operator can in principle be embedded within any metaheuristic, we selected HGS due to its flexibility, ability to effectively solve multi-period problems with time constraints, and strong empirical performance across various VRP settings \citep{vidal2013hybrid, vidal2014unified}. In addition to its successful application in prior research, HGS has also demonstrated top-tier performance in recent benchmarking competitions, such as the DIMACS VRP Challenge and the EURO/NeurIPS Vehicle Routing Competition \citep{Kool2023}, where it formed the core of leading solution methods. These results further support the robustness of HGS as a general-purpose search framework for complex VRPs.
}

\begin{figure}[htbp]
\vspace*{-0.5cm}
\hspace*{-1cm}
\includegraphics[width=1.2\textwidth]{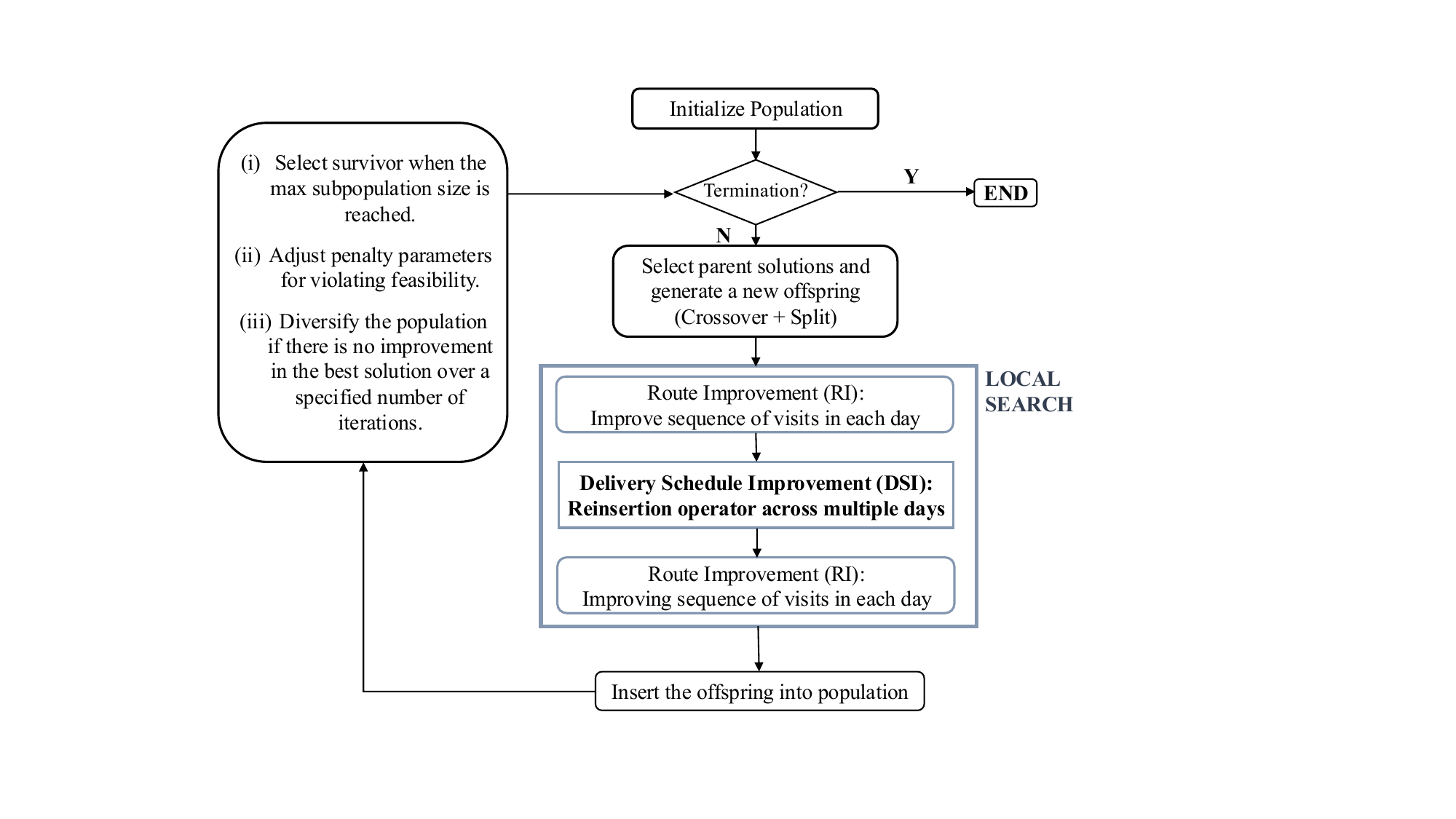}\vspace*{-1cm}
\caption{High-level description of the HGS-IRP algorithm}
\label{HGS}
\end{figure}

Figure~\ref{HGS} outlines the HGS-IRP framework, which follows the structure of the original HGS but includes several key adaptations. The solution is represented by a collection of \textit{giant tours}—one per day—that encode customer sequences. Combined with delivery quantities, they define a complete IRP solution. The population includes both feasible and infeasible individuals, with penalized capacity violations encouraged to foster diversification.

After generating an initial population (Section~\ref{hgs:is}), the algorithm iteratively applies selection and crossover (Section~\ref{hgs:psc}), followed by route decoding via the Split algorithm \citep{prins2004simple}. A local search phase (Section~\ref{hgs:edu}) then refines each offspring through (i) route improvement (RI) while keeping delivery quantities fixed, (ii) delivery schedule improvement (DSI) using the DS operator, and (iii) re-optimization of routes (RI).
After educating the offspring, the algorithm inserts the resulting solution in the correct subpopulation according to its feasibility and possibly repairs it to additionally include the resulting solution in the feasible subpopulation. If the maximum size of any subpopulation is reached, a survivor selection process is triggered to reduce this subpopulation size (Section~\ref{hgs:ss}). Subsequently, the penalty parameter associated with capacity violations is adjusted. If the best solution does not improve within a predefined number of iterations, the population is diversified by introducing new solutions generated through the same procedure as in the initialization.

\subsection{Solution Representation.}
\label{hgs:sr}

Each individual in the genetic algorithm represents a solution. It contains, on each day, the sequence of visits to retailers without trip delimiters, i.e., represented as a ``giant tour'' \citep{prins2004simple}. This representation is advantageous since it is possible to optimally split this giant tour to create individual routes using a linear-time split algorithm \citep{Vidal2016}. For the IRP, additional information is needed to completely represent solutions: the quantities delivered to each retailer on each day. Therefore, each solution includes, on each day, (i) a \textit{giant tour chromosome} encoding the visit sequence and (ii) a \textit{quantity chromosome} specifying the corresponding delivery quantities.

\subsection{Generation of Initial Solutions}
\label{hgs:is}

Each initial solution is constructed using a just-in-time delivery rule. Starting from day $t = 1$, a retailer is scheduled for a visit if its stock is insufficient to meet demand for the next period. The delivery quantity is determined by the order-up-to level (OU) policy, meaning the retailer's inventory is replenished up to its maximum capacity $U_i$. Additionally, even if a retailer has sufficient inventory, there is a 30\% probability of making a delivery anyway, still up to reaching level $U_i$. This process continues until the final day $H$. Once the set of daily quantities are fixed, the visits are ordered randomly on each day, partitioned into vehicle routes using the Split algorithm, and the education operator (Section~\ref{hgs:edu}) is applied to refine the solution.

\subsection{Selection and Crossover}
\label{hgs:psc}

Each parent is selected by a binary tournament within the union of the feasible and infeasible subpopulations. The crossover procedure begins by randomly shuffling the days and selecting two random indices $j_1$ and $j_2$ with uniform probability such that $j_1 < j_2$ from this list. These indices divide the days into three sets.

For days with indices strictly below $j_1$, a segment of deliveries is inherited from Parent 1.
This segment is determined by selecting a random start and end position within the parent’s sequence for that day, as in the PIX crossover of \cite{vidal2012hybrid}. The corresponding deliveries, along with their respective delivery quantities, are transferred to the offspring. On days with indices within $\{j_1,\dots,j_2-1\}$, no deliveries are inherited. Finally, for the remaining days, the entire delivery sequence from Parent 1 is copied to the offspring.

Once inheritance from Parent 1 is complete, missing deliveries are filled using Parent 2. Specifically, for days with indices strictly below $j_2$, where no sequence or only a partial sequence was inherited from Parent 1, additional deliveries from Parent 2 are introduced. Each delivery is added at the end of the sequence only if it does not already appear on that day. The corresponding delivery quantities are also inherited, unless they would exceed the retailer's inventory constraints, in which case the maximum feasible delivery quantity is selected.

After inheritance is finalized, the split algorithm is applied on each day to partition each giant tour into individual routes. The resulting solution is then refined through the local-search-based education process described in the next section.

\subsection{Education}
\label{hgs:edu}

The local-search education procedure improves solutions by reorganizing the delivery schedule across all routes and days. This operation is applied to complete solutions. The process follows three sequential steps: RI-DSI-RI. The first and last steps, referred to as Route Improvement (RI), involve a standard local search procedure identical to that of \cite{vidal2012hybrid}, based on nine classic neighborhood operators, including \textsc{Relocate} and \textsc{Swap} (involving up to two consecutive nodes), as well as \textsc{2-Opt} and \textsc{2-Opt*}. These neighborhoods are explored in a randomized order, with any improving move immediately accepted until a local minimum is reached.
For the Delivery Schedule Improvement (DSI) phase,  as described in Section~\ref{edu:DSopt}, \textcolor{black}{ we iterate over all retailers in random order to optimize their delivery schedules by reassigning visit days. 
This phase stops when a complete pass over all retailers results in no improvement in any retailer’s delivery plan.}

\subsection{Diversity Management}
\label{hgs:ss}

When a subpopulation reaches its maximum size, the algorithm employs a survivor selection process, similar to the method proposed in \cite{vidal2012hybrid}, to determine which individuals are retained for the next generation. This mechanism aims to maintain diversity in visit patterns while preserving elite solutions based on cost. Consequently, discarded individuals are either clones or solutions that perform poorly in terms of both cost and contribution to population diversity.

To quantify diversity, a distance measure specific to the IRP setting is used. Given two individuals, $P_1$ and $P_2$, their distance is computed as:
\begin{equation}
\Delta(P_1,P_2) = \frac{1}{n} \sum_{i=1}^n \mathds{1} (\pi_i(P_1) \neq \pi_i(P_2))
\end{equation}
where $\mathds{1} (\pi_i(P_1) \neq \pi_i(P_2))$ equals $1$ if the set of delivery days $\pi_i(P_1)$ assigned to retailer $i$ in $P_1$ differs from that in $P_2$, and $0$ otherwise. 

As in \cite{vidal2012hybrid}, this distance measure is used to define a biased fitness function that influences selection probability, ensuring that the survivor selection process effectively maintains diversity while eliminating redundant or low-quality solutions.

\section{Computational experiments} 
\label{Section:comp_experiments}

This section presents the computational experiments conducted to evaluate the performance of the HGS algorithm using the new DS operator for the IRP. We first focus on the classical IRP without stock-out, as it is a  benchmark in the literature, and the results are detailed in Section~\ref{bench}. 
We then investigate the empirical behavior of the number of linear pieces in the cost-to-go functions \(C_t\), introduced in Section~\ref{sec:complexity}, to assess the practical computational complexity of the DS operator. The corresponding analysis and results are presented in Section~\ref{sec:complexity-experiment}.
Finally, we evaluate the algorithm's performance on the IRP with stock-out, providing further insights into the trade-offs between customer service levels and delivery costs in Section~\ref{Perf}.\\

\noindent
\textbf{Benchmark Instances.}
We test the algorithm on established IRP benchmark instances, categorized into \textit{small} and \textit{large} instance datasets. The \textit{small} dataset consists of well-known instances from \citep{archetti2007branch}, while the \textit{large} instances were introduced in \cite{archetti2012hybrid}.

Each instance in the \textit{small dataset} is defined by three parameters: the planning horizon ($H$), the number of retailers ($n$), and the inventory cost level ($C$). The dataset is divided into four categories, covering two values of $H$ ($H = 3$ and $H = 6$) and two levels of inventory cost: low inventory cost (LC) and high inventory cost (HC). The number of retailers $n$ in each category ranges from 10 to 50. Each problem size includes five distinct instances.
The number of available vehicles $K$ varies between 2 and 5, resulting in a total of 800 instances in the small dataset.
The \textit{large dataset} introduced in \cite{archetti2012hybrid} follows a similar structure, with $H = 6$ and $n \in \{50, 100, 200\}$. The settings for $C$ and $K$ remain consistent with the small instances. For each of the 24 possible parameter combinations ($n$, $H$, $C$, $K$), 10 randomly generated instances were created, leading to a total of 240 large instances.
More recently, \cite{skaalnes2024new} introduced a new set of IRP instances designed to be more challenging than the benchmark instances described above. We refer to the small and large datasets as \textit{benchmark instances}, while we denote the instances from \cite{skaalnes2024new} as \textit{SAHS instances} (named as the initials of the authors). In Section~\ref{bench}, we evaluate the HGS algorithm on SAHS instances as well. These instances feature varying numbers of retailers (10 to 200), planning horizons of 6, 9, or 12 periods, and multiple vehicle types with capacities of 8, 18, or 38 EUR-pallets. Unlike the benchmark dataset, which assumes constant retailer demand, SAHS instances incorporate variable demand over time to better simulate real-world conditions.

For the tests in Section \ref{Perf}, where stock-out is allowed, we used the small benchmark dataset and set the value of the stock-out penalty $\rho$ as follows: 50,  100, 300, and 1,000,000, the latter being set to a large value to avoid stock-out. Thus, we obtained \textcolor{black}{four} instances for each original IRP instance.\\

\noindent
\textbf{Experimental Setup.}
All experiments were conducted on a server equipped with Intel(R) Xeon(R) Gold 6336Y CPUs running at 2.40~GHz. 
Each experiment was executed on a single CPU. The open-source code and 
instances are available at~\url{https://github.com/vidalt/HGS-IRP/}, along with solution details for each instance.

The HGS algorithm terminates when any of the following stopping criteria are met: 
(i) a total of 100,000 iterations, 
(ii) 10,000 consecutive iterations without improvement, or 
(iii) a maximum runtime of 2,400 seconds for small-scale instances and 7,200 seconds for large-scale and SAHS instances. This configuration permits the use of a computational time, on large instances, commensurate to that used by the other methods.

\subsection{Performance analysis on the IRP}
\label{bench}

Classical benchmark instances for the IRP do not permit stock-out. To accommodate this setting, we omit negative $\hat I_i^t$ in the DP transition processes described in Section~\ref{edu:DSopt} while retaining the rest of the algorithm unchanged. We compare the performance of our algorithm, HGS-IRP, against the most recent heuristics in the literature as well as the best-performing method from the DIMACS 2022 Implementation Challenge on Vehicle Routing, which included a dedicated track on the IRP. All benchmark instances and reference solutions used in our experiments are sourced from \url{https://or-brescia.unibs.it/instances} and \url{http://dimacs.rutgers.edu/programs/challenge/vrp/irp/}, and are also provided in the Git repository associated with this project. HGS-IRP is evaluated against three leading methods: the unified decomposition matheuristic proposed by \citet{chitsaz2019unified} (denoted CCJ), the kernel search heuristic of \citet{archetti2021kernel} (KS), and the Branch-and-Heuristic approach developed by \citet{skaalnes2023branch} (BCHeu), which achieved first place in the DIMACS competition. This selection provides a comprehensive basis for performance comparison.

\begin{table}
\centering
\caption{\textcolor{black}{Comparative analysis on medium-scale IRP instances}}
\label{tab:MIRP}
\resizebox{1\textwidth}{!}{
\begin{tabular}{cccrrrcrrrlrrrlrrrrrcr} 
\toprule
                    &                      &                      & \multicolumn{3}{c}{CCJ}                                                          & \multicolumn{1}{r}{} & \multicolumn{3}{c}{KS}                                                           & \multicolumn{1}{c}{} & \multicolumn{3}{c}{BCHeu}                                                        & \multicolumn{1}{r}{} & \multicolumn{7}{c}{HGS-IRP}                                                                                                                                                                        \\ 
\cmidrule{4-6}\cmidrule{8-10}\cmidrule{12-14}\cmidrule{16-20}
$H$                 & $C$                  & $n$                  & \multicolumn{1}{c}{Sol} & \multicolumn{1}{c}{Gap(\%)} & \multicolumn{1}{c}{T(s)} &                      & \multicolumn{1}{c}{Sol} & \multicolumn{1}{c}{Gap(\%)} & \multicolumn{1}{c}{T(s)} & \multicolumn{1}{c}{} & \multicolumn{1}{c}{Sol} & \multicolumn{1}{c}{Gap(\%)} & \multicolumn{1}{c}{T(s)} & \multicolumn{1}{c}{} & \multicolumn{1}{c}{Avg} & \multicolumn{1}{c}{$\mathrm{Gap}_\mathrm{A}$(\%)}        & \multicolumn{1}{c}{T(s)} & \multicolumn{1}{c}{Best} & \multicolumn{1}{c}{$\mathrm{Gap}_\mathrm{B}$(\%)} &                      & \multicolumn{1}{c}{BKS}  \\ 
\midrule
\multirow{22}{*}{3} & \multirow{11}{*}{HC} & 5                    & 2855.05                 & 3.39                        & 1.7                      &                      & 2761.47                 & 0.00                        & 5.9                      &                      & 2761.47                 & 0.00                        & 7200.0                   &                      & 2761.47                 & 0.00                               & 60.3                     & 2761.47                  & 0.00                        &                      & 2761.47                  \\
                    &                      & 10                   & 5013.81                 & 3.24                        & 4.3                      &                      & 4856.89                 & 0.01                        & 91.3                     &                      & 4856.50                 & 0.00                        & 7200.0                   &                      & 4856.50                 & 0.00                               & 104.3                    & 4856.50                  & 0.00                        &                      & 4856.50                  \\
                    &                      & 15                   & 5620.86                 & 1.63                        & 11.3                     &                      & 5534.01                 & 0.06                        & 114.9                    &                      & 5530.87                 & 0.00                        & 7200.0                   &                      & 5530.88                 & 0.00                               & 119.6                    & 5530.87                  & 0.00                        &                      & 5530.87                  \\
                    &                      & 20                   & 7191.44                 & 1.39                        & 25.3                     &                      & 7105.34                 & 0.18                        & 202.6                    &                      & 7093.86                 & 0.02                        & 7200.0                   &                      & 7092.57                 & 0.00                               & 127.0                    & 7092.57                  & 0.00                        &                      & 7092.57                  \\
                    &                      & 25                   & 8501.63                 & 0.86                        & 37.6                     &                      & 8438.19                 & 0.10                        & 324.3                    &                      & 8429.51                 & 0.00                        & 7200.0                   &                      & 8430.00                 & 0.01                               & 153.9                    & 8429.50                  & 0.00                        &                      & 8429.42                  \\
                    &                      & 30                   & 9699.83                 & 0.40                        & 64.1                     &                      & 9721.92                 & 0.63                        & 504.8                    &                      & 9665.34                 & 0.04                        & 7200.0                   &                      & 9660.65                 & 0.00                               & 207.5                    & \textbf{9660.57}         & \textbf{-0.01}              &                      & 9661.11                  \\
                    &                      & 35                   & 10210.21                & 0.39                        & 85.2                     &                      & 10205.92                & 0.35                        & 544.2                    &                      & 10172.89                & 0.02                        & 7200.0                   &                      & 10171.15                & 0.01                               & 259.5                    & 10170.59                 & 0.00                        &                      & 10170.59                 \\
                    &                      & 40                   & 10986.12                & 0.46                        & 141.9                    &                      & 11012.14                & 0.70                        & 614.7                    &                      & 10945.29                & 0.09                        & 7200.0                   &                      & 10950.63                & 0.13                               & 318.9                    & \textbf{10935.93}        & 0.00                        &                      & 10935.94                 \\
                    &                      & 45                   & 11998.55                & 0.27                        & 179.9                    &                      & 12077.35                & 0.93                        & 663.1                    &                      & 11999.87                & 0.28                        & 7200.0                   &                      & 11976.60                & 0.09                               & 313.1                    & 11966.17                 & 0.00                        &                      & 11966.14                 \\
                    &                      & 50                   & 13571.12                & 0.33                        & 232.1                    &                      & 13692.82                & 1.23                        & 1173.8                   &                      & 13546.34                & 0.15                        & 7200.0                   &                      & 13528.67                & 0.02                               & 453.9                    & \textbf{13523.46}        & \textbf{-0.02}              &                      & 13526.27                 \\
                    &                      & Avg                  & 8564.86                 & 0.85                        & 78.3                     &                      & 8540.60                 & 0.56                        & 424.0                    &                      & 8500.19                 & 0.08                        & 7200.0                   &                      & 8495.91                 & 0.03                               & 211.8                    & \textbf{8492.76}         & 0.00                        &                      & 8493.09                  \\ 
\cmidrule{2-22}
                    & \multirow{11}{*}{LC} & 5                    & 2180.27                 & 4.81                        & 1.7                      &                      & 2080.31                 & 0.00                        & 6.2                      &                      & 2080.31                 & 0.00                        & 7200.0                   &                      & 2080.32                 & 0.00                               & 60.4                     & 2080.31                  & 0.00                        &                      & 2080.31                  \\
                    &                      & 10                   & 3221.39                 & 5.51                        & 4.2                      &                      & 3053.18                 & 0.00                        & 89.3                     &                      & 3053.14                 & 0.00                        & 7200.0                   &                      & 3053.13                 & 0.00                               & 99.4                     & 3053.13                  & 0.00                        &                      & 3053.13                  \\
                    &                      & 15                   & 3200.77                 & 2.66                        & 10.3                     &                      & 3121.36                 & 0.11                        & 118.5                    &                      & 3118.09                 & 0.01                        & 7200.0                   &                      & 3117.85                 & 0.00                               & 107.6                    & 3117.85                  & 0.00                        &                      & 3117.85                  \\
                    &                      & 20                   & 3771.04                 & 2.27                        & 20.5                     &                      & 3702.66                 & 0.41                        & 225.8                    &                      & 3687.81                 & 0.01                        & 7200.0                   &                      & 3687.49                 & 0.00                               & 128.6                    & 3687.48                  & 0.00                        &                      & 3687.48                  \\
                    &                      & 25                   & 3994.91                 & 1.49                        & 60.6                     &                      & 3951.99                 & 0.40                        & 337.7                    &                      & 3937.48                 & 0.03                        & 7200.0                   &                      & 3938.14                 & 0.04                               & 149.6                    & 3936.56                  & 0.00                        &                      & 3936.37                  \\
                    &                      & 30                   & 3989.68                 & 1.07                        & 44.1                     &                      & 3978.59                 & 0.78                        & 482.1                    &                      & 3953.57                 & 0.15                        & 7200.0                   &                      & 3947.76                 & 0.00                               & 192.3                    & 3947.63                  & 0.00                        &                      & 3947.63                  \\
                    &                      & 35                   & 4237.04                 & 0.94                        & 58.1                     &                      & 4224.87                 & 0.65                        & 556.9                    &                      & 4198.92                 & 0.03                        & 7200.0                   &                      & 4197.83                 & 0.01                               & 207.2                    & 4197.62                  & 0.00                        &                      & 4197.58                  \\
                    &                      & 40                   & 4397.24                 & 1.89                        & 94.4                     &                      & 4377.20                 & 1.43                        & 619.9                    &                      & 4328.67                 & 0.30                        & 7200.0                   &                      & 4322.94                 & 0.17                               & 264.5                    & \textbf{4315.28}         & \textbf{-0.01}              &                      & 4315.63                  \\
                    &                      & 45                   & 4517.73                 & 1.03                        & 107.4                    &                      & 4576.23                 & 2.33                        & 698.3                    &                      & 4491.29                 & 0.44                        & 7200.0                   &                      & 4473.40                 & 0.04                               & 274.6                    & 4471.83                  & 0.00                        &                      & 4471.83                  \\
                    &                      & 50                   & 5296.46                 & 1.16                        & 152.7                    &                      & 5445.02                 & 4.00                        & 1098.2                   &                      & 5253.17                 & 0.34                        & 7200.0                   &                      & 5239.09                 & 0.07                               & 365.5                    & \textbf{5232.78}         & \textbf{-0.05}              &                      & 5235.53                  \\
                    &                      & Avg                  & 3880.65                 & 2.01                        & 55.4                     &                      & 3851.14                 & 1.23                        & 423.3                    &                      & 3810.24                 & 0.16                        & 7200.0                   &                      & 3805.79                 & 0.04                               & 185.0                    & \textbf{3798.08}         & \textbf{-0.16}              &                      & 3804.33                  \\ 
\cmidrule{2-22}
\multirow{22}{*}{6} & \multirow{11}{*}{HC} & 5                    & –                       & –                           & –                        &                      & 7371.68                 & 0.00                        & 95.6                     &                      & 7371.44                 & 0.00                        & 7200.0                   &                      & \textbf{7090.34}        & \textbf{-3.81}                     & 394.3                    & \textbf{7090.08}         & \textbf{-3.82}              &                      & 7371.44                  \\
                    &                      & 10                   & 11760.57                & 7.54                        & 13.1                     &                      & 10954.19                & 0.17                        & 134.1                    &                      & 10936.89                & 0.01                        & 7200.0                   &                      & 10940.78                & 0.05                               & 828.9                    & 10935.72                 & 0.00                        &                      & 10935.67                 \\
                    &                      & 15                   & 13797.25                & 5.31                        & 77.7                     &                      & 13156.04                & 0.42                        & 198.5                    &                      & 13110.17                & 0.07                        & 7200.0                   &                      & 13117.52                & 0.12                               & 1042.8                   & 13105.75                 & 0.03                        &                      & 13101.33                 \\
                    &                      & 20                   & 16979.09                & 4.03                        & 71.5                     &                      & 16420.79                & 0.61                        & 503.7                    &                      & 16334.12                & 0.08                        & 7200.0                   &                      & 16335.86                & 0.09                               & 1358.5                   & \textbf{16320.20}        & 0.00                        &                      & 16320.60                 \\
                    &                      & 25                   & 19081.64                & 3.97                        & 135.5                    &                      & 18498.31                & 0.79                        & 694.3                    &                      & 18395.07                & 0.23                        & 7200.0                   &                      & 18376.31                & 0.13                               & 1590.9                   & 18356.10                 & 0.02                        &                      & 18353.33                 \\
                    &                      & 30                   & 21314.31                & 2.41                        & 221.3                    &                      & 21034.43                & 1.06                        & 687.2                    &                      & 20866.21                & 0.25                        & 7200.0                   &                      & 20836.78                & 0.11                               & 1766.2                   & \textbf{20812.69}        & 0.00                        &                      & 20813.61                 \\
                    &                      & 35                   & –                       & –                           & –                        &                      & –                       & –                           & –                        &                      & 22671.70                & 0.29                        & 7200.0                   &                      & 22640.50                & 0.15                               & \textbf{1888.2}          & 22609.81                 & 0.01                        &                      & 22606.89                 \\
                    &                      & 40                   & –                       & –                           & –                        &                      & –                       & –                           & –                        &                      & 24292.25                & 0.63                        & 7200.0                   &                      & \textbf{23451.28}       & \textbf{-2.86}                     & \textbf{1923.8}          & \textbf{23400.59}        & \textbf{-3.07}              &                      & 24141.23                 \\
                    &                      & 45                   & –                       & –                           & –                        &                      & –                       & –                           & –                        &                      & 26722.37                & 0.21                        & 7200.0                   &                      & \textbf{25920.28}       & \textbf{-2.80}                     & \textbf{2007.8}          & \textbf{25880.87}        & \textbf{-2.95}              &                      & 26667.58                 \\
                    &                      & 50                   & –                       & –                           & –                        &                      & –                       & –                           & –                        &                      & 30467.00                & 0.15                        & 7200.0                   &                      & \textbf{29567.18}       & \textbf{-2.81}                     & \textbf{2201.4}          & \textbf{29515.75}        & \textbf{-2.97}              &                      & 30420.51                 \\
                    &                      & Avg                  & –                       & –                           & –                        &                      & –                       & –                           & –                        &                      & 19116.72                & 0.23                        & 7200.0                   &                      & \textbf{18827.68}       & \textbf{-1.16}                     & \textbf{1500.27}         & \textbf{18802.75}        & \textbf{-1.42}              &                      & 19073.22                 \\ 
\cmidrule{2-22}
                    & \multirow{11}{*}{LC} & 5                    & 5794.88                 & 5.78                        & 5.8                      &                      & 5478.02                 & 0.00                        & 88.7                     &                      & 5477.99                 & 0.00                        & 7200.0                   &                      & 5478.03                 & 0.00                               & 428.7                    & 5477.99                  & 0.00                        &                      & 5477.99                  \\
                    &                      & 10                   & 8213.51                 & 9.15                        & 11.0                     &                      & 7548.02                 & 0.31                        & 130.0                    &                      & 7525.02                 & 0.00                        & 7200.0                   &                      & 7526.81                 & 0.03                               & 725.4                    & 7525.61                  & 0.01                        &                      & 7524.83                  \\
                    &                      & 15                   & 8469.11                 & 7.79                        & 27.2                     &                      & 7891.16                 & 0.44                        & 220.5                    &                      & 7863.14                 & 0.08                        & 7200.0                   &                      & 7869.35                 & 0.16                               & 897.9                    & 7858.85                  & 0.03                        &                      & 7856.69                  \\
                    &                      & 20                   & 10123.49                & 6.94                        & 59.1                     &                      & 9558.74                 & 0.97                        & 641.7                    &                      & 9506.25                 & 0.42                        & 7200.0                   &                      & 9488.26                 & 0.23                               & 1260.5                   & 9476.98                  & 0.11                        &                      & 9466.51                  \\
                    &                      & 25                   & 10648.25                & 5.86                        & 108.4                    &                      & 10227.55                & 1.68                        & 723.7                    &                      & 10104.99                & 0.46                        & 7200.0                   &                      & 10088.21                & 0.29                               & 1525.1                   & 10064.89                 & 0.06                        &                      & 10058.85                 \\
                    &                      & 30                   & 10394.63                & 4.84                        & 153.9                    &                      & 10222.89                & 3.10                        & 619.1                    &                      & 9971.14                 & 0.57                        & 7200.0                   &                      & 9933.83                 & 0.19                               & 1735.6                   & \textbf{9901.21}         & \textbf{-0.14}              &                      & 9915.06                  \\
                    &                      & 35                   & –                       & –                           & –                        &                      & –                       & –                           & –                        &                      & 10915.93                & 0.79                        & 7200.0                   &                      & 10843.66                & 0.12                               & \textbf{1925.4}          & \textbf{10817.43}        & \textbf{-0.12}              &                      & 10830.43                 \\
                    &                      & 40                   & –                       & –                           & –                        &                      & –                       & –                           & –                        &                      & 11002.22                & 0.27                        & 7200.0                   &                      & 10991.20                & 0.17                               & \textbf{1985.1}          & \textbf{10951.25}        & \textbf{-0.20}              &                      & 10972.65                 \\
                    &                      & 45                   & –                       & –                           & –                        &                      & –                       & –                           & –                        &                      & 11579.88                & 0.46                        & 7200.0                   &                      & 11544.00                & 0.15                               & \textbf{2025.8}          & \textbf{11510.08}        & \textbf{-0.15}              &                      & 11527.10                 \\
                    &                      & 50                   & –                       & –                           & –                        &                      & –                       & –                           & –                        &                      & 13511.62                & 0.26                        & 7200.0                   &                      & 13505.18                & 0.21                               & \textbf{2262.0}          & \textbf{13454.90}        & \textbf{-0.16}              &                      & 13476.69                 \\
                    &                      & Avg                  & –                       & –                           & –                        &                      & –                       & –                           & –                        &                      & 9745.82                 & 0.36                        & 7200.0                   &                      & 9726.85                 & 0.17                               & \textbf{1477.2}          & \textbf{9703.92}         & \textbf{-0.07}              &                      & 9710.68                  \\ 
\cmidrule{2-22}
\multirow{8}{*}{6}  & \multirow{4}{*}{LC}  & 50                   & 14336.56                & 4.96                        & 394.5                    &                      & 14455.92                & 5.84                        & 2589.5                   &                      & 13694.50                & 0.26                        & 7200.0                   &                      & 13670.93                & 0.09                               & 4512.1                   & \textbf{13619.39}        & \textbf{-0.29}              &                      & 13658.58                 \\
                    &                      & 100                  & 17630.85                & 4.50                        & 1823.2                   &                      & 18421.73                & 9.18                        & 2112.0                   &                      & 16890.54                & 0.11                        & 7200.0                   &                      & \textbf{16805.40}       & \textbf{-0.40}                     & 6391.5                   & \textbf{16736.78}        & \textbf{-0.80}              &                      & 16872.38                 \\
                    &                      & 200                  & 25142.35                & 4.12                        & 8310.1                   &                      & 27786.21                & 15.07                       & 828.3                    &                      & 24146.93                & 0.00                        & 7200.0                   &                      & 24216.63                & 0.29                               & 7140.9                   & \textbf{24063.84}        & \textbf{-0.34}              &                      & 24146.93                 \\
                    &                      & Avg                  & 19036.59                & 4.45                        & 3509.3                   &                      & 20221.29                & 10.95                       & 1843.3                   &                      & 18243.99                & 0.10                        & 7200.0                   &                      & 18230.99                & 0.03                               & 6014.8                   & \textbf{18140.00}        & \textbf{-0.47}              &                      & 18225.96                 \\ 
\cmidrule{2-22}
                    & \multirow{4}{*}{HC}  & 50                   & 31283.24                & 2.15                        & 642.5                    &                      & 31412.44                & 2.57                        & 3081.1                   &                      & 30663.33                & 0.13                        & 7200.0                   &                      & 30649.56                & 0.08                               & 4323.8                   & \textbf{30581.14}        & \textbf{-0.14}              &                      & 30624.92                 \\
                    &                      & 100                  & 52530.23                & 1.30                        & 2968.6                   &                      & 53591.81                & 3.35                        & 2360.4                   &                      & 51865.07                & 0.02                        & 7200.0                   &                      & 51888.67                & 0.06                               & 5531.8                   & \textbf{51740.12}        & \textbf{-0.22}              &                      & 51856.76                 \\
                    &                      & 200                  & 96076.10                & 0.82                        & 16435.8                  &                      & 99299.77                & 4.20                        & 1507.6                   &                      & 95298.01                & 0.00                        & 7200.0                   &                      & 95576.34                & 0.29                               & 6927.1                   & 95358.71                 & 0.06                        &                      & 95298.01                 \\
                    &                      & Avg                  & 59963.19                & 1.19                        & 6682.3                   &                      & 61434.67                & 3.67                        & 2316.4                   &                      & 59275.47                & 0.03                        & 7200.0                   &                      & 59371.52                & 0.19                               & 5594.2                   & \textbf{59226.66}        & \textbf{-0.06}              &                      & 59259.90                 \\ 
\midrule
Avg Gap             & \multicolumn{1}{l}{} & \multicolumn{1}{l}{} & \multicolumn{1}{c}{}    & \multicolumn{1}{c}{2.84}    &                          & \multicolumn{1}{r}{} &                         & \multicolumn{1}{c}{2.66}    &                          & \multicolumn{1}{r}{} &                         & \multicolumn{1}{l}{0.17}    &                          & \multicolumn{1}{r}{} &                         & \multicolumn{1}{l}{\textbf{-0.15}} &                          &                          &                       \textbf{-0.38}      & \multicolumn{1}{r}{} &                          \\
Avg Time            & \multicolumn{1}{l}{} & \multicolumn{1}{l}{} & \multicolumn{1}{c}{}    & \multicolumn{1}{c}{}        & 1509.1                   & \multicolumn{1}{r}{} &                         & \multicolumn{1}{c}{}        & 879.7                    & \multicolumn{1}{r}{} &                         & \multicolumn{1}{l}{}        & 7200.0                   & \multicolumn{1}{r}{} &                         & \multicolumn{1}{l}{}               & 1989.4                   &                          &                             & \multicolumn{1}{r}{} &                          \\
\bottomrule
\end{tabular}
}
\end{table}
Results are summarized in Table~\ref{tab:MIRP}, which reports average values for instances sharing the same values of $n$, $H$, and $C$, on the benchmark instances, small and large (the latter corresponding to the last two blocks). For each method, we report the solution value (Sol), the percentage gap (Gap) relative to the best known solution (BKS, shown in the last column), and the computational time in seconds (T(s)). The gap is computed as $\left(\frac{\mathrm{Sol}}{\mathrm{BKS}} - 1\right) \times 100\%$. To the best of our knowledge, the BKS values are compiled from all existing methods for the IRP under an ML-based inventory policy, including the best-performing solutions reported in the DIMACS 2022 Vehicle Routing Implementation Challenge.
 In contrast to these single-run results, we report the average results of our method over multiple runs, as this provides a more reliable performance estimator given the random nature of the algorithm. 
 \textcolor{black}{Specifically, we present the average solution value (Avg), average gap ($\mathrm{Gap}_\mathrm{A}$) to the best known solution (BKS), average CPU time (T(s)), the best solution  (Best) among the 10 runs and its corresponding gap to the BKS ($\mathrm{Gap}_\mathrm{B}$). }

The proposed HGS-IRP algorithm exhibits strong performance in both solution quality and computational efficiency. It consistently produces solutions that are comparable to, or better than, those obtained by existing methods, all within reasonable computational times. Across all 1,040 tested instances, the maximum optimality gap relative to the BKS is only 0.29\%, while the average gap is $-0.15\%$, both of which outperform all benchmark algorithms.
 In terms of computational time, HGS-IRP is particularly efficient on large-scale instances, and it also maintains strong solution quality on smaller instances. As the problem size increases, both in number of periods ($H$) and customers ($n$), the advantages of HGS-IRP become more pronounced. On the small 3-day dataset (400 instances), HGS-IRP identifies 37 new best-known solutions, matches the existing BKS in 320 instances, and performs slightly worse in only 43 instances. For the small 6-day dataset, it finds 150 new best solutions and matches the BKS in 83 instances. Most notably, on the large 6-day dataset, HGS-IRP finds new best solutions in 178 out of 240 instances, representing a substantial improvement over the current state-of-the-art in solving large-scale IRP problems.

\begin{result} 
HGS-IRP yields 768 best-known solutions, out of which 365 are new best solutions, across 1040 benchmark IRP instances. In detail, for small instances, HGS-IRP identifies \textbf{187 new best solutions} out of 800 instances. For large instances, HGS-IRP finds \textbf{178 new best solutions} out of 240 instances. This highlights the efficiency and effectiveness of HGS-IRP, especially in handling larger and more complex datasets, and demonstrates the effectiveness of the proposed DS operator, which is the only operator in charge of optimizing delivery schedules.
\end{result}

\noindent
\textbf{Experiments on SAHS instances.}
\label{newdata}
We further evaluate HGS-IRP  on the challenging IRP instances recently introduced by \cite{skaalnes2024new}. Table~\ref{tab:new-data} presents the average results across instances of varying sizes.
\cite{skaalnes2024new} propose three approaches, and we report the corresponding results: (1) Construction heuristic,
(2) Matheuristic,
(3) Branch-and-Cut.
Only the Branch-and-Cut method provides lower bounds (LB), in contrast to the constructive and matheuristic methods, which are primarily designed to generate feasible solutions and, therefore, only provide upper bounds (UB).
Finally, BKS reports the best of the feasible solutions found by the three approaches (the minimum value in UB).

\begin{result}
The computational results demonstrate that, out of a total of 270 benchmark instances, the best-known solution is improved in \textbf{191} instances and matched in an additional \textbf{25} instances. Notably, within the same 2-hour time limit, substantially better solutions are obtained, \textcolor{black}{yielding an average improvement of 18.21\% over all instances compared to the BKS previously reported by \cite{skaalnes2024new}}.
\end{result}

\begin{table}
\centering
\caption{\textcolor{black}{Comparative analysis on large-scale IRP instances}}
\label{tab:new-data}
\resizebox{1\textwidth}{!}{%
\begin{tabular}{cccclccclcccclcccccc} 
\toprule
    & \multicolumn{3}{c}{Construction} & \multicolumn{1}{c}{} & \multicolumn{3}{c}{Matheuristic} & \multicolumn{1}{c}{} & \multicolumn{4}{c}{Branch and Cut}      & \multicolumn{1}{c}{} & \multicolumn{5}{c}{HGS-IRP}                                                           & \multicolumn{1}{l}{}  \\ 
\cmidrule(l){2-4}\cmidrule{6-8}\cmidrule{10-13}\cmidrule{15-19}
$n$ & UB        & Gap(\%) & T(s)       &                      & UB       & Gap(\%) & T(s)        &                      & LB       & UB        & Gap(\%) & T(s)   &                      & Avg                & $\mathrm{Gap}_\mathrm{A}$(\%)         & T(s)    & Best               & $\mathrm{Gap}_\mathrm{B}$(\%)        & BKS                   \\ 
\midrule
10  & 7502.74   & 40.08   & 0.0        &                      & 5385.52  & 0.55    & 1748.8      &                      & 5331.14  & 5356.03   & 0.00    & 5422.3 &                      & 5369.76            & 0.26            & 4378.4  & 5356.81            & 0.01            & 5356.03               \\
25  & 19431.74  & 40.37   & 0.0        &                      & 15072.20 & 8.88    & 1800.0      &                      & 13626.19 & 13842.88  & 0.00    & 7199.1 &                      & 13886.24           & 0.31            & 7078.1  & \textbf{13819.07}  & \textbf{-0.17}  & 13842.88              \\
50  & 42969.72  & 6.84    & 0.1        &                      & -        & -       & 2400.0      &                      & 28862.31 & 51389.61  & 27.78   & 7193.2 &                      & \textbf{29945.03}  & \textbf{-25.54} & 7390.6  & \textbf{29753.66}  & \textbf{-26.02} & 40218.66              \\
100 & 86098.23  & 0.38    & 0.7        &                      & -        & -       & 2400.0      &                      & 52307.74 & 172930.13 & 101.61  & 7193.8 &                      & \textbf{58382.45}  & \textbf{-31.94} & 7215.3  & \textbf{58121.23}  & \textbf{-32.24} & 85775.08              \\
200 & 154146.38 & 0.00    & 6.7        &                      & -        & -       & 2400.0      &                      & 76747.91 & -         & -       & 7217.9 &                      & \textbf{106730.50} & \textbf{-30.76} & 11139.3 & \textbf{105188.88} & \textbf{-31.76} & 154146.38             \\
\bottomrule
\end{tabular}
}
\end{table}

\subsection{Empirical Behavior of the Number of Pieces in $C_t$}
\label{sec:complexity-experiment}

In Section~\ref{sec:complexity}, we analyzed the theoretical complexity of computing the cost-to-go functions $C_t$. The number of linear pieces in $C_t$ is theoretically bounded by $O(U_i)$, and the most expensive operation is the convolution of $C_{t-1}$ and $F_t$, which takes $O(U_i K)$ time. When repeating across all~$H$ time steps of the dynamic program, the total complexity of evaluating the DS neighborhood for a single retailer becomes $O(U_i K H)$. However, in practice, we observe that the number of pieces remains much smaller than this bound.

\begin{figure}[htbp]
    \centering
    \includegraphics[width=1\linewidth]{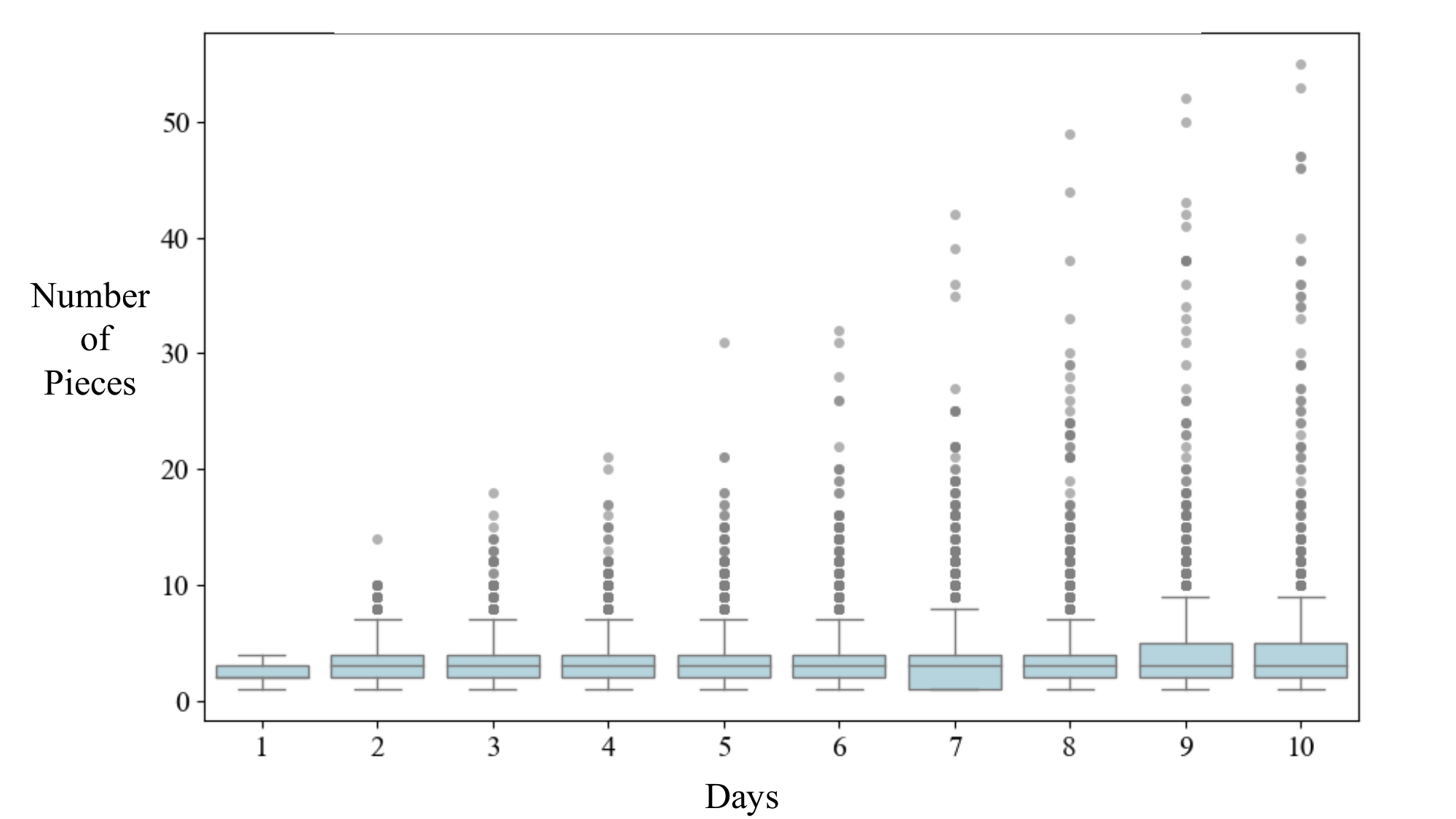}
    \caption{Boxplots of the number of pieces in $C_t$ from Day 1 to Day 10. Each box represents the distribution across different instances. The box spans from the 25th to the 75th percentile, with the median shown as a horizontal line.}
    \label{fig:piece_boxplot}
\end{figure}

To evaluate this, we recorded the number of pieces in $C_t$ for several test instances.  The results are shown in Figure~\ref{fig:piece_boxplot}. These results were obtained on an instance from the small instance set, which includes 50 customers and  high holding cost. Each day, up to $K = 5$ vehicles are available for deliveries, and the maximum inventory level $U_i$ varies significantly across customers, ranging from 22 to 294. The planning horizon in this instance is $H = 10$ days. These parameters define the theoretical bounds on the number of linear pieces in $C_t$, as discussed in Section~\ref{sec:complexity}, where the full evaluation of the DS neighborhood for one retailer requires $O(U_i K H)$ time.
We began recording from iteration 1001 to 1020, after running 1000 iterations to allow the population to stabilize.
During this period, measurements were taken only after crossover, and the results of the first application of the DS operator were recorded within the DSI loop during the local search.
We collected the number of linear pieces in $C_t$ for each of the 50 customers on 20 DSI calls, resulting in 1000 data points per day (50 customers $\times$ 20 rounds) for the boxplot. We repeated the same procedure on four other instances with varying horizon lengths ($H \in \{6, 10, 12\}$), and observed similar trends. Therefore, the behavior shown in Figure~\ref{fig:piece_boxplot} shows consistent patterns across different time horizons. Although it may seem surprising at first, the number of pieces initially increases but then stabilizes or even decreases over time, as the algorithm retains only the most promising options and prunes the rest. Moreover, while the upper range tends to grow in later days, the median number of pieces remains low, and reductions are frequently observed. This suggests that the actual computational burden remains well controlled.


\textcolor{black}{Two key design mechanisms in our algorithm explain this observation:}

\begin{enumerate}

    \item \textbf{Range-based pruning}: The function $C_t(I_i^t)$ models the cost of ending day $t$ with inventory level $I_i^t$. As established in Section~\ref{sec:complexity}, the feasible domain for $I_i^t$ is $[0, U_i - d_i^t]$, where $U_i$ is the maximum inventory and $d_i^t$ is the demand on day $t$. This domain applies to both transitions~\eqref{eq:metho1} and~\eqref{eq:metho2} in the DP recursion. Before performing convolutions or minimizations, we discard any segments of the piecewise linear functions that fall entirely outside this interval. Additionally, segments that do not cover any integer-valued inventory levels are also removed, as they contain no feasible solutions. These two steps significantly reduce the number of pieces carried forward at each stage.

    \item \textbf{Optimality-based merging}: When merging cost functions from different transitions (e.g., delivery vs. no-delivery,  different stock-out levels), we retain only the segment with the lowest cost over each interval. Dominated segments are discarded and they are frequently encountered in Equations~\eqref{eq:metho3} and~\eqref{eq:metho4}. This operation further limits the total number of pieces and avoids redundant computations.
\end{enumerate}

Together, these two mechanisms effectively limit the number of function segments. Despite the theoretical worst-case scenario, the number of linear segments in $C_t$ remains small over the planning horizon, leading to an efficient approach.

\subsection{Performance of HGS-IRP on IRP with penalized stock-outs}
\label{Perf}

In the stock-out-allowed setting, no standard benchmark instances are available. Therefore, we evaluate the performance of HGS-IRP on both the small and large benchmark IRP datasets described earlier. Specifically, we test four different values of the stock-out penalty parameter $\rho$: 50, 100, 300, and 1,000,000.

\begin{figure}[htbp]
\centering
\begin{overpic}[width=1\textwidth]{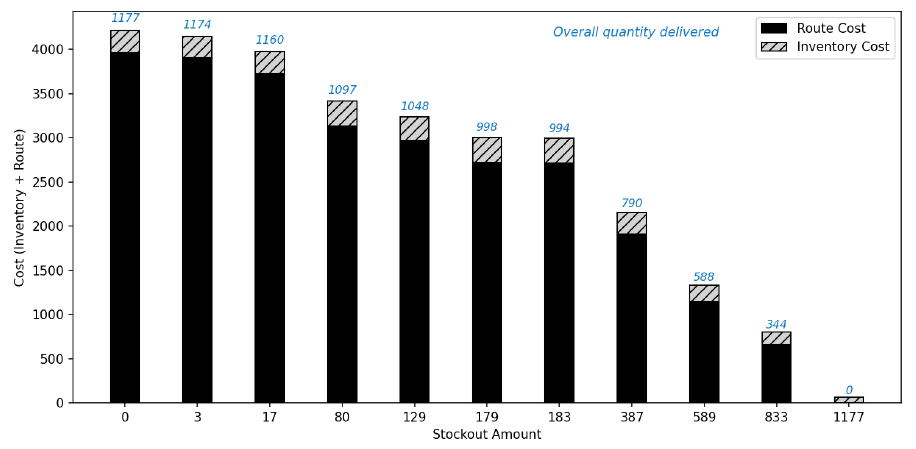}
\end{overpic}
\caption{{\footnotesize Trade-off between stock-out amount and routing/inventory cost for a specific instance (15 customers and 5 vehicles) under varying stock-out penalties $\rho$, which varies \textcolor{black}{from 1 to 300}  in integer increments. }}
\label{fig:sto-in}
\end{figure}

Figure~\ref{fig:sto-in} illustrates the trade-off observed on a representative instance when varying the penalty parameter $\rho$. 
The $x$-axis shows the stock-out quantity, and the total quantity delivered to retailers is displayed on top of each column. 
The $y$-axis represents the sum of the routing and inventory costs, with routing costs shown in black and inventory costs shown as shaded segments.
\textcolor{black}{Each bar also reports the total quantity delivered to all customers, which decreases as the stock-out amount increases.}
The plot shows that lower values of $\rho$ result in higher stock-out quantities and reduced total cost. In particular, routing costs decrease consistently as stock-outs increase, as the algorithm avoids serving some retailers when the penalty is low. This leads to a shorter or less frequent delivery schedule. Conversely, as $\rho$ increases, the algorithm favors full delivery, reducing stock-outs but increasing overall cost due to more extensive routing and higher inventory levels. 
Note that multiple values of $\rho$ can yield the same best delivery decision. As a result, several distinct $\rho$ values may correspond to a single bar in the figure, with identical stock-out quantities and cost decompositions.
It is observed that once $\rho$ exceeds 91, no stock-out events occur. Consequently, for $\rho \geq 92$, all corresponding bars in the figure coincide with the first bar, as increasing the penalty further only reinforces stock-out avoidance but does not change the outcome.

\textcolor{black}{Table~\ref{tab:final} also reports detailed results for all tested instances under the four values of $\rho$. Instances are grouped by number of retailers, inventory cost type (HC and LC), and time horizon. For each configuration, we present the best solution value, the average solution value over 10 runs, and the corresponding computational time.  These results confirm a general trend: higher $\rho$ values typically result in increased total cost and runtime. We additionally report the gap  (Gap) in average cost across $\rho$ values, using the $\rho=1,000,000$ case as a benchmark which approximates the no-stock-out case. This highlights the impact of stock-out penalties on the overall cost structure. As expected, higher $\rho$ values yield higher total costs, as the routing plans adjust to avoid larger stock-out penalties. Under the HC (high holding cost) setting, cost differences across $\rho$ are smaller, since high holding and stock-out costs already discourage stock-outs. By contrast, in the LC setting, small $\rho$ values produce relatively low routing-cost solutions, as stock-outs remain cheaper than replenishment.}
Overall, HGS-IRP maintains stable performance across the range of penalty settings, suggesting robustness in handling various stock-out scenarios.

\begin{table}[htbp]
\centering
\caption{Best results (Best), average results (Avg),  average runtime in seconds (T(s)) on instances of varying sizes for different stock-out penalty settings.}
\resizebox{1\textwidth}{!}{%
\begin{tabular}{ccrrrrrrrrrrrrrrr} 
\toprule
\multirow{3}{*}{\begin{tabular}[c]{@{}c@{}}Data\\Set\end{tabular}}   & \multirow{3}{*}{$n$} & \multicolumn{4}{c}{$\rho = 50$}                                                                             & \multicolumn{4}{c}{$\rho = 100$}                                                                            & \multicolumn{4}{c}{$\rho = 300$}                                                                            & \multicolumn{3}{c}{$\rho = 1,000,000$}                                         \\ 
\cmidrule{3-17}
                                                                     &                      & \multicolumn{1}{c}{Best} & \multicolumn{3}{c}{Avg}                                                          & \multicolumn{1}{c}{Best} & \multicolumn{3}{c}{Avg}                                                          & \multicolumn{1}{c}{Best} & \multicolumn{3}{c}{Avg}                                                          & \multicolumn{1}{c}{Best} & \multicolumn{2}{c}{Avg}                             \\ 
\cmidrule(r){3-3}\cmidrule(r){4-6}\cmidrule(lr){7-7}\cmidrule(lr){8-10}\cmidrule(r){11-11}\cmidrule(r){12-15}\cmidrule(lr){16-17}
                                                                     &                      & \multicolumn{1}{c}{Best} & \multicolumn{1}{c}{T(s)} & \multicolumn{1}{c}{Avg} & \multicolumn{1}{c}{Gap(\%)} & \multicolumn{1}{c}{Best} & \multicolumn{1}{c}{T(s)} & \multicolumn{1}{c}{Avg} & \multicolumn{1}{c}{Gap(\%)} & \multicolumn{1}{c}{Best} & \multicolumn{1}{c}{T(s)} & \multicolumn{1}{c}{Avg} & \multicolumn{1}{c}{Gap(\%)} & \multicolumn{1}{c}{Best} & \multicolumn{1}{c}{T(s)} & \multicolumn{1}{c}{Avg}  \\ 
\midrule
\multirow{11}{*}{\begin{tabular}[c]{@{}c@{}}LC\\~$H$=3\end{tabular}} & 5                    & 587.08                   & 16.9                     & 587.08                  & -71.78                      & 948.97                   & 24.8                     & 948.97                  & -54.38                      & 1674.99                  & 48.9                     & 1674.99                 & -19.48                      & 2080.31                  & 87.1                     & 2080.31                  \\
                                                                     & 10                   & 1497.86                  & 44.5                     & 1497.86                 & -50.94                      & 2264.81                  & 59.7                     & 2264.81                 & -25.82                      & 2955.57                  & 108.0                    & 2955.57                 & -3.20                       & 3053.13                  & 139.0                    & 3053.13                  \\
                                                                     & 15                   & 1921.61                  & 68.7                     & 1921.61                 & -38.37                      & 2540.49                  & 87.7                     & 2540.49                 & -18.52                      & 3031.66                  & 130.9                    & 3031.97                 & -2.75                       & 3117.85                  & 155.4                    & 3117.85                  \\
                                                                     & 20                   & 2553.68                  & 88.6                     & 2553.68                 & -30.75                      & 3212.50                  & 121.1                    & 3212.50                 & -12.88                      & 3611.01                  & 162.5                    & 3611.01                 & -2.07                       & 3687.48                  & 184.5                    & 3687.49                  \\
                                                                     & 25                   & 3124.38                  & 115.2                    & 3126.72                 & -20.64                      & 3643.10                  & 159.4                    & 3643.22                 & -7.53                       & 3909.73                  & 211.2                    & 3914.53                 & -0.64                       & 3936.38                  & 215.9                    & 3939.80                  \\
                                                                     & 30                   & 3353.37                  & 149.2                    & 3353.42                 & -15.05                      & 3777.17                  & 198.3                    & 3777.38                 & -4.32                       & 3920.38                  & 253.3                    & 3921.61                 & -0.66                       & 3947.63                  & 274.8                    & 3947.74                  \\
                                                                     & 35                   & 3664.29                  & 163.8                    & 3668.62                 & -12.60                      & 4088.36                  & 239.6                    & 4089.93                 & -2.57                       & 4187.51                  & 286.1                    & 4187.64                 & -0.24                       & 4197.61                  & 297.4                    & 4197.70                  \\
                                                                     & 40                   & 3782.98                  & 205.9                    & 3795.75                 & -12.20                      & 4102.82                  & 262.3                    & 4107.49                 & -4.99                       & 4271.98                  & 339.1                    & 4272.86                 & -1.17                       & 4315.37                  & 387.7                    & 4323.25                  \\
                                                                     & 45                   & 4022.20                  & 227.4                    & 4028.15                 & -9.97                       & 4293.90                  & 293.0                    & 4299.02                 & -3.92                       & 4458.13                  & 361.2                    & 4463.54                 & -0.24                       & 4471.83                  & 386.8                    & 4474.28                  \\
                                                                     & 50                   & 4676.77                  & 262.4                    & 4684.09                 & -10.59                      & 5020.13                  & 381.7                    & 5033.01                 & -3.93                       & 5181.11                  & 485.6                    & 5191.22                 & -0.91                       & 5232.76                  & 539.9                    & 5239.03                  \\
                                                                     & Avg                  & 2918.42                  & 134.2                    & 2921.70                 & -23.24                      & 3389.23                  & 182.8                    & 3391.68                 & -10.89                      & 3720.21                  & 238.7                    & 3722.49                 & -2.20                       & 3804.04                  & 266.8                    & 3806.06                  \\ 
\midrule
\multirow{11}{*}{\begin{tabular}[c]{@{}c@{}}LC~\\$H$=6\end{tabular}} & 5                    & 2114.42                  & 84.3                     & 2114.42                 & -64.15                      & 3318.14                  & 174.0                    & 3318.15                 & -43.75                      & 5192.22                  & 372.3                    & 5192.30                 & -11.98                      & 5898.73                  & 621.4                    & 5898.77                  \\
                                                                     & 10                   & 3944.96                  & 169.9                    & 3944.96                 & -47.59                      & 5732.51                  & 324.8                    & 5732.56                 & -23.84                      & 7302.65                  & 797.4                    & 7304.50                 & -2.95                       & 7524.83                  & 1016.5                   & 7526.86                  \\
                                                                     & 15                   & 5469.63                  & 310.0                    & 5469.87                 & -30.48                      & 6937.18                  & 618.4                    & 6938.69                 & -11.82                      & 7728.22                  & 1028.9                   & 7734.11                 & -1.71                       & 7858.97                  & 1189.9                   & 7868.57                  \\
                                                                     & 20                   & 7038.84                  & 470.3                    & 7040.12                 & -25.78                      & 8654.39                  & 973.2                    & 8660.29                 & -8.71                       & 9370.41                  & 1563.2                   & 9390.04                 & -1.01                       & 9475.69                  & 1680.2                   & 9486.06                  \\
                                                                     & 25                   & 8044.29                  & 733.5                    & 8048.87                 & -20.23                      & 9382.32                  & 1414.2                   & 9402.07                 & -6.82                       & 9983.37                  & 1770.7                   & 10006.16                & -0.84                       & 10063.19                 & 1865.5                   & 10090.52                 \\
                                                                     & 30                   & 8635.57                  & 1187.4                   & 8644.47                 & -13.02                      & 9628.74                  & 1707.7                   & 9646.79                 & -2.94                       & 9902.33                  & 2046.3                   & 9932.44                 & -0.06                       & 9910.42                  & 2072.6                   & 9938.70                  \\
                                                                     & 35                   & 9594.55                  & 1470.9                   & 9617.88                 & -11.37                      & 10512.11                 & 1937.7                   & 10541.70                & -2.86                       & 10813.82                 & 2142.5                   & 10841.46                & -0.09                       & 10819.22                 & 2182.5                   & 10851.70                 \\
                                                                     & 40                   & 9695.89                  & 1640.3                   & 9750.94                 & -11.38                      & 10597.52                 & 2053.2                   & 10634.63                & -3.35                       & 10940.88                 & 2202.3                   & 10988.62                & -0.13                       & 10953.99                 & 2225.9                   & 11003.23                 \\
                                                                     & 45                   & 10441.67                 & 1951.6                   & 10466.43                & -9.36                       & 11176.42                 & 2138.1                   & 11209.45                & -2.92                       & 11500.24                 & 2248.4                   & 11542.45                & -0.04                       & 11509.90                 & 2262.3                   & 11547.06                 \\
                                                                     & 50                   & 12418.36                 & 2141.8                   & 12474.57                & -7.78                       & 13171.02                 & 2279.9                   & 13224.56                & -2.23                       & 13420.02                 & 2366.0                   & 13484.94                & -0.31                       & 13460.64                 & 2368.9                   & 13526.78                 \\
                                                                     & Avg                  & 7739.82                  & 1016.0                   & 7757.25                 & -20.63                      & 8911.04                  & 1362.1                   & 8930.89                 & -8.62                       & 9615.42                  & 1653.8                   & 9641.70                 & -1.35                       & 9747.56                  & 1748.6                   & 9773.82                  \\ 
\midrule
\multirow{11}{*}{\begin{tabular}[c]{@{}c@{}}HC~\\$H$=3\end{tabular}} & 5                    & 2635.88                  & 63.7                     & 2635.88                 & -4.55                       & 2733.03                  & 81.2                     & 2733.48                 & -1.01                       & 2754.16                  & 83.1                     & 2754.16                 & -0.26                       & 2761.47                  & 83.7                     & 2761.47                  \\
                                                                     & 10                   & 4829.84                  & 125.3                    & 4830.24                 & -0.54                       & 4846.23                  & 139.8                    & 4846.28                 & -0.21                       & 4856.50                  & 145.6                    & 4856.56                 & 0.00                        & 4856.50                  & 143.1                    & 4856.58                  \\
                                                                     & 15                   & 5491.10                  & 144.4                    & 5491.72                 & -0.71                       & 5509.68                  & 159.7                    & 5509.68                 & -0.38                       & 5524.66                  & 165.1                    & 5524.72                 & -0.11                       & 5530.87                  & 170.2                    & 5530.88                  \\
                                                                     & 20                   & 7066.76                  & 170.8                    & 7067.11                 & -0.36                       & 7082.51                  & 180.1                    & 7082.51                 & -0.14                       & 7090.65                  & 183.2                    & 7090.65                 & -0.03                       & 7092.57                  & 181.9                    & 7092.57                  \\
                                                                     & 25                   & 8421.57                  & 204.3                    & 8421.96                 & -0.10                       & 8428.26                  & 215.7                    & 8428.45                 & -0.02                       & 8429.46                  & 228.0                    & 8429.98                 & 0.00                        & 8429.50                  & 221.9                    & 8429.97                  \\
                                                                     & 30                   & 9656.95                  & 270.2                    & 9657.25                 & -0.04                       & 9658.73                  & 293.2                    & 9658.84                 & -0.02                       & 9660.57                  & 301.5                    & 9660.59                 & 0.00                        & 9660.57                  & 295.7                    & 9660.69                  \\
                                                                     & 35                   & 10166.57                 & 348.7                    & 10167.74                & -0.03                       & 10167.88                 & 378.4                    & 10168.28                & -0.03                       & 10170.59                 & 369.8                    & 10171.19                & 0.00                        & 10170.63                 & 360.3                    & 10171.22                 \\
                                                                     & 40                   & 10919.59                 & 408.2                    & 10933.05                & -0.17                       & 10932.19                 & 438.6                    & 10952.13                & 0.01                        & 10934.26                 & 463.1                    & 10953.13                & 0.02                        & 10935.95                 & 462.8                    & 10951.37                 \\
                                                                     & 45                   & 11958.89                 & 391.3                    & 11964.93                & -0.04                       & 11961.90                 & 419.3                    & 11965.66                & -0.04                       & 11966.25                 & 430.9                    & 11977.74                & 0.06                        & 11966.17                 & 420.6                    & 11970.11                 \\
                                                                     & 50                   & 13519.58                 & 573.5                    & 13529.78                & -0.03                       & 13522.07                 & 635.8                    & 13531.65                & -0.02                       & 13523.54                 & 639.6                    & 13530.39                & -0.03                       & 13523.45                 & 609.7                    & 13534.32                 \\
                                                                     & Avg                  & 8466.67                  & 270.0                    & 8469.97                 & -0.31                       & 8484.25                  & 294.2                    & 8487.70                 & -0.10                       & 8491.06                  & 301.0                    & 8494.91                 & -0.01                       & 8492.77                  & 295.0                    & 8495.92                  \\ 
\midrule
\multirow{11}{*}{\begin{tabular}[c]{@{}c@{}}HC~\\$H$=6\end{tabular}} & 5                    & 7313.49                  & 487.1                    & 7313.82                 & -0.79                       & 7352.60                  & 570.9                    & 7352.92                 & -0.26                       & 7367.91                  & 643.0                    & 7368.72                 & -0.05                       & 7371.85                  & 601.3                    & 7372.14                  \\
                                                                     & 10                   & 10849.74                 & 911.9                    & 10854.04                & -0.80                       & 10905.93                 & 1041.9                   & 10910.54                & -0.28                       & 10931.97                 & 1069.5                   & 10937.92                & -0.03                       & 10935.99                 & 1060.2                   & 10941.57                 \\
                                                                     & 15                   & 13066.26                 & 1256.9                   & 13074.88                & -0.32                       & 13104.46                 & 1297.8                   & 13116.45                & -0.01                       & 13106.68                 & 1396.8                   & 13117.23                & 0.00                        & 13106.99                 & 1323.3                   & 13117.17                 \\
                                                                     & 20                   & 16307.98                 & 1634.5                   & 16321.56                & -0.09                       & 16319.36                 & 1692.4                   & 16335.14                & -0.01                       & 16321.38                 & 1736.8                   & 16336.75                & 0.00                        & 16322.96                 & 1688.1                   & 16336.76                 \\
                                                                     & 25                   & 18320.33                 & 1905.2                   & 18350.60                & -0.18                       & 18340.80                 & 1842.6                   & 18370.67                & -0.07                       & 18358.87                 & 1884.7                   & 18382.77                & 0.00                        & 18358.73                 & 1865.3                   & 18383.15                 \\
                                                                     & 30                   & 20810.58                 & 2016.5                   & 20835.51                & 0.01                        & 20810.16                 & 2040.6                   & 20837.67                & 0.02                        & 20808.01                 & 2088.0                   & 20836.47                & 0.02                        & 20805.62                 & 2028.9                   & 20833.26                 \\
                                                                     & 35                   & 22607.50                 & 2103.9                   & 22643.27                & 0.02                        & 22607.88                 & 2161.8                   & 22644.03                & 0.02                        & 22614.14                 & 2149.1                   & 22644.51                & 0.02                        & 22606.38                 & 2072.9                   & 22639.84                 \\
                                                                     & 40                   & 24146.43                 & 2029.4                   & 24203.13                & 0.07                        & 24144.45                 & 2127.1                   & 24227.38                & 0.17                        & 24143.91                 & 2111.3                   & 24224.99                & 0.16                        & 24147.37                 & 2179.5                   & 24186.13                 \\
                                                                     & 45                   & 27129.80                 & 617.0                    & 27129.80                & 0.22                        & 27129.80                 & 617.0                    & 27129.80                & 0.22                        & 27070.40                 & 559.0                    & 27165.30                & 0.35                        & 27036.50                 & 2400.0                   & 27071.16                 \\
                                                                     & 50                   & 30397.51                 & 2282.3                   & 30491.50                & 0.10                        & 30413.11                 & 2277.1                   & 30518.17                & 0.19                        & 30425.21                 & 2275.0                   & 30497.33                & 0.12                        & 30405.67                 & 2361.1                   & 30460.95                 \\
                                                                     & Avg                  & 19094.96                 & 1524.5                   & 19121.81                & -0.06                       & 19112.85                 & 1566.9                   & 19144.28                & 0.05                        & 19114.85                 & 1591.3                   & 19151.20                & 0.09                        & 19109.81                 & 1758.0                   & 19134.21                 \\ 
\midrule
\multirow{4}{*}{HC}                                                  & 50                   & 30615.71                 & 7200.0                   & 30697.45                & -0.05                       & 30619.20                 & 7200.0                   & 30696.18                & -0.06                       & 30618.82                 & 7200.0                   & 30699.39                & -0.05                       & 30629.73                 & 7200.0                   & 30713.76                 \\
                                                                     & 100                  & 51806.43                 & 7200.2                   & 51970.67                & -0.01                       & 51806.30                 & 7200.2                   & 51959.48                & -0.03                       & 51803.57                 & 7200.2                   & 51953.48                & -0.05                       & 51803.26                 & 7200.2                   & 51976.93                 \\
                                                                     & 200                  & 95497.49                 & 7200.9                   & 95754.60                & 0.00                        & 95523.94                 & 7201.0                   & 95763.75                & 0.01                        & 95534.84                 & 7201.0                   & 95753.63                & 0.00                        & 95526.64                 & 7201.0                   & 95757.36                 \\
                                                                     & Avg                  & 59306.54                 & 7200.4                   & 59474.24                & -0.01                       & 59316.48                 & 7200.4                   & 59473.14                & -0.02                       & 59319.08                 & 7200.4                   & 59468.83                & -0.02                       & 59319.88                 & 7200.4                   & 59482.68                 \\ 
\midrule
\multirow{4}{*}{LC}                                                  & 50                   & 12551.19                 & 7200.0                   & 12615.56                & -8.06                       & 13336.96                 & 7200.0                   & 13405.90                & -2.30                       & 13623.71                 & 7200.0                   & 13689.28                & -0.23                       & 13660.14                 & 7200.0                   & 13721.51                 \\
                                                                     & 100                  & 16341.00                 & 7200.1                   & 16444.51                & -2.51                       & 16656.30                 & 7200.2                   & 16754.31                & -0.68                       & 16783.32                 & 7200.2                   & 16867.82                & 0.00                        & 16787.39                 & 7200.2                   & 16868.29                 \\
                                                                     & 200                  & 23910.12                 & 7200.9                   & 24110.79                & -1.11                       & 24135.77                 & 7200.9                   & 24347.91                & -0.14                       & 24183.88                 & 7201.0                   & 24389.36                & 0.03                        & 24176.11                 & 7200.9                   & 24382.03                 \\
                                                                     & Avg                  & 17600.77                 & 7200.4                   & 17723.62                & -3.28                       & 18043.01                 & 7200.4                   & 18169.37                & -0.84                       & 18196.97                 & 7200.4                   & 18315.49                & -0.05                       & 18207.88                 & 7200.4                   & 18323.94                 \\
\bottomrule\end{tabular}}
\label{tab:final}
\end{table}

\section{Conclusions} 
\label{Section:Conclusions}

In this paper, we have introduced a novel large neighborhood search operator for the Inventory Routing Problem (IRP). For any given retailer, the operator removes all visits across the planning horizon and computes an optimized delivery schedule ---reselecting visit days and delivery quantities to jointly minimize routing and inventory costs. In contrast to existing search operators, which either apply local changes to the delivery schedule or incur high computational costs when considering broader modifications, this is the first operator specifically designed for the IRP that efficiently optimizes replenishment decisions over the entire planning horizon. The operator also handles stock-outs in a natural way, a feature that has received limited attention in the IRP literature despite being common in practical settings.

The operator is implemented through a dynamic programming (DP) procedure, using piecewise-linear functions and supported by preprocessing and acceleration techniques to ensure practical efficiency.  \textcolor{black}{It is designed to be metaheuristic-independent: it only requires an incumbent solution and can therefore be embedded as a large-scale neighborhood within various frameworks, such as Iterated Local Search (ILS), adaptive Large Neighborhood Search (LNS), or other metaheuristics. }
As demonstrated in our experiments, this design enables the calculation of reinsertion schedules within microseconds, making it useful for systematic solution refinement within metaheuristics. We integrated the operator into the Hybrid Genetic Search (HGS) framework, where it serves as the sole component responsible for modifying delivery schedules and quantities. The resulting HGS-IRP algorithm achieves state-of-the-art performance: it discovers new best-known solutions for 365 out of 1,040 benchmark instances and improves the best feasible solutions for 191 out of 270 challenging larger-scale IRP instances introduced in 2023.

The research perspectives stemming from this work are numerous. 
\textcolor{black}{While the integration of the DS operator into HGS already yields good performance, the operator itself is generic and can be embedded into other methods more specifically tailored to the IRP. Moreover, it could be extended to richer IRP variants, such as those with time windows, heterogeneous fleets, multiple depots, or to the production routing problem, by adapting the evaluation function to incorporate the corresponding constraints and cost components.}
Moreover, this search strategy could be adapted to other deterministic IRP variants, as well as to more complex settings such as two-stage stochastic IRPs and dynamic IRPs, which are increasingly relevant in real-world applications. In these contexts, the operator could be adapted to solve sample average approximation-based deterministic approximations of stochastic problems, either as a standalone approach within a rolling horizon framework or as a generator of solutions for training imitation learning models, such as the one developed in \cite{greif2024combinatorial}.

\section*{Acknowledgments}
\textcolor{black}{The authors thank the three anonymous reviewers whose comments helped to improve and clarify this manuscript. They also} thank Dr. Khalid Laaziri for his assistance in IT-related matters, including but not limited to resolving server issues and problems with Python and Matplotlib, and Dr. Brett Kraabel for help with writing.
\textcolor{black}{This work was supported by SCALE-AI through its Research Chairs program,  the National Natural Science Foundation of China (72571234), and the SRIBD Grant (J00220250002).}


%

%
%


\bibliographystyle{ormsv080} 
\bibliography{tddarp.bib} 

@inproceedings{Kool2023,
author = {Kool, W. and Bliek, L. and Numeroso, D. and Zhang, Y. and Catshoek, T. and Tierney, K. and Vidal, T. and Gromicho, J.},
booktitle = {Proceedings of the NeurIPS 2022 Competitions Track},
pages = {35--49},
series = {PMLR},
title = {{The EURO Meets NeurIPS 2022 vehicle routing competition}},
volume = {220},
year = {2022}
}

@article{abdelmaguid2006genetic,
 author = {Abdelmaguid, T.F. and Dessouky, M.M.},
 journal = {International Journal of Production Research},
 number = {21},
 pages = {4445--4464},
 publisher = {Taylor \& Francis},
 title = {A genetic algorithm approach to the integrated inventory-distribution problem},
 volume = {44},
 year = {2006}
}

@article{dinh2025matheuristic,
  title={Matheuristic Algorithms for the Inventory Routing Problem With Unsplit and Split Deliveries},
  author={Dinh, N. M. and Archetti, C. and Bertazzi, L.},
  journal={Networks},
  year={2025},
  publisher={Wiley Online Library}
}

@article{CoelhoLaporte2013,
author = {Coelho, L. C. and Laporte, G.},
title = {A branch-and-cut algorithm for the multi-product multi-vehicle inventory-routing problem},
journal = {International Journal of Production Research},
volume = {51},
number = {23-24},
pages = {7156-7169},
year  = {2013},
publisher = {Taylor \& Francis}
}

@article{CoelhoLaporte2013b,
title = {The exact solution of several classes of inventory-routing problems},
author = {Coelho, L. C. and Laporte, G.},
journal = {Computers \& Operations Research},
volume = {40},
number = {2},
pages = {558-565},
year = {2013},
issn = {0305-0548}
}

@article{Archetti2014,
Author = {Archetti, C. and Bianchessi, N. and Irnich, S. and Speranza, M. G.},
title = {Formulations for an inventory routing problem.},
journal = {International Transactions in Operational Research},
Pages = {353--374},
volume = {21},
number = {3},
year = {2014}
}

@article{CoelhoLaporte2014,
title = {Improved solutions for inventory-routing problems through valid inequalities and input ordering},
author = {Coelho, L. C. and Laporte, G.},
journal = {International Journal of Production Economics},
volume = {155},
pages = {391 - 397},
year = {2014},
issn = {0925-5273}
}

@article{DesaulniersRC2016,
title = {A branch-price-and-cut algorithm for the Inventory Routing Problem},
author = {Desaulniers, G. and Rakke, J. G. and Coelho, L. C.},
journal = {Transportation Science},
volume = {50},
number = {3},
pages = {1060-1076},
year = {2016}
}

@article{adulyasak2014formulations,
 author = {Adulyasak, Y. and Cordeau, J. and Jans, R.},
 journal = {INFORMS Journal on Computing},
 number = {1},
 pages = {103--120},
 publisher = {INFORMS},
 title = {Formulations and branch-and-cut algorithms for multivehicle production and inventory routing problems},
 volume = {26},
 year = {2014}
}

@article{adulyasak2014optimization,
 author = {Adulyasak, Y. and Cordeau, J. and Jans, R.},
 journal = {Transportation Science},
 number = {1},
 pages = {20--45},
 publisher = {INFORMS},
 title = {Optimization-based adaptive large neighborhood search for the production routing problem},
 volume = {48},
 year = {2014}
}

@article{ahuja2002survey,
 author = {Ahuja, R.K. and Ergun, Ö. and Orlin, J.B. and Punnen, A.P.},
 journal = {Discrete Applied Mathematics},
 number = {1-3},
 pages = {75--102},
 publisher = {Elsevier},
 title = {A survey of very large-scale neighborhood search techniques},
 volume = {123},
 year = {2002}
}

@article{Bouvier2024,  
author = {Bouvier, L. and Dalle, G. and Parmentier, A. and Vidal, T.},   
journal = {Transportation Science},
number = {1},
pages = {131--151},
title = {{Solving a continent-scale inventory routing problem at Renault}}, 
volume = {58},
year = {2024}
}

@article{alvarez2018iterated,
 author = {Alvarez, A. and Munari, P. and Morabito, R.},
 journal = {International Transactions in Operational Research},
 number = {6},
 pages = {1785--1809},
 publisher = {Wiley Online Library},
 title = {Iterated local search and simulated annealing algorithms for the inventory routing problem},
 volume = {25},
 year = {2018}
}

@article{andersson2010industrial,
 author = {Andersson, H. and Hoff, A. and Christiansen, M. and Hasle, G. and L{\o}kketangen, A.},
 journal = {Computers \& operations research},
 number = {9},
 pages = {1515--1536},
 publisher = {Elsevier},
 title = {Industrial aspects and literature survey: Combined inventory management and routing},
 volume = {37},
 year = {2010}
}

@article{applegate2003traveling,
 author = {Applegate, D. and Bixby, R.E. and Chvátal, V. and Cook, W.},
 journal = {SIAM Journal on Computing},
 number = {1},
 pages = {1--8},
 publisher = {SIAM},
 title = {A new bound on the traveling salesman problem},
 volume = {32},
 year = {2003}
}

@article{archetti2007branch,
 author = {Archetti, C. and Bertazzi, L. and Laporte, G. and Speranza, M.G.},
 journal = {Transportation science},
 number = {3},
 pages = {382--391},
 publisher = {INFORMS},
 title = {A branch-and-cut algorithm for a vendor-managed inventory-routing problem},
 volume = {41},
 year = {2007}
}

@article{archetti2012hybrid,
 author = {Archetti, C. and Bertazzi, L. and Hertz, A. and Speranza, M.G.},
 journal = {INFORMS Journal on Computing},
 number = {1},
 pages = {101--116},
 publisher = {INFORMS},
 title = {A hybrid heuristic for an inventory routing problem},
 volume = {24},
 year = {2012}
}

@article{archetti2017matheuristic,
 author = {Archetti, C. and Boland, N. and Grazia Speranza, M.},
 journal = {INFORMS Journal on Computing},
 number = {3},
 pages = {377--387},
 publisher = {INFORMS},
 title = {A matheuristic for the multivehicle inventory routing problem},
 volume = {29},
 year = {2017}
}

@article{archetti2021kernel,
 author = {Archetti, C. and Guastaroba, G. and Huerta-Mu{\~n}oz, D.L. and Speranza, M.G.},
 journal = {International Transactions in Operational Research},
 number = {6},
 pages = {2984--3013},
 publisher = {Wiley Online Library},
 title = {A kernel search heuristic for the multivehicle inventory routing problem},
 volume = {28},
 year = {2021}
}

@article{bell1983improving,
 author = {Bell, W.J. and Dalberto, L.M. and Fisher, M.L. and Greenfield, A.J. and Jaikumar, R. and Kedia, P. and Mack, R.G. and Prutzman, P.J.},
 journal = {Interfaces},
 number = {6},
 pages = {4--23},
 publisher = {INFORMS},
 title = {Improving the distribution of industrial gases with an on-line computerized routing and scheduling optimizer},
 volume = {13},
 year = {1983}
}

@article{bouvier2024solving,
 author = {Bouvier, L. and Dalle, G. and Parmentier, A. and Vidal, T.},
 journal = {Transportation Science},
 number = {1},
 pages = {131--151},
 publisher = {Informs},
 title = {Solving a continent-scale inventory routing problem at renault},
 volume = {58},
 year = {2024}
}

@article{campbell2004decomposition,
 author = {Campbell, A.M. and Savelsbergh, M.W.},
 journal = {Transportation science},
 number = {4},
 pages = {488--502},
 publisher = {INFORMS},
 title = {A decomposition approach for the inventory-routing problem},
 volume = {38},
 year = {2004}
}

@article{carter1996solving,
 author = {Carter, M.W. and Farvolden, J.M. and Laporte, G. and Xu, J.},
 journal = {INFOR: Information Systems and Operational Research},
 number = {4},
 pages = {290--306},
 publisher = {Taylor \& Francis},
 title = {Solving an integrated logistics problem arising in grocery distribution},
 volume = {34},
 year = {1996}
}

@article{chitsaz2019unified,
 author = {Chitsaz, M. and Cordeau, J. and Jans, R.},
 journal = {INFORMS Journal on Computing},
 number = {1},
 pages = {134--152},
 publisher = {INFORMS},
 title = {A unified decomposition matheuristic for assembly, production, and inventory routing},
 volume = {31},
 year = {2019}
}

@article{christiaens2020slack,
 author = {Christiaens, J. and Vanden Berghe, G.},
 journal = {Transportation Science},
 number = {2},
 pages = {417--433},
 publisher = {INFORMS},
 title = {Slack induction by string removals for vehicle routing problems},
 volume = {54},
 year = {2020}
}

@article{christiansen1999decomposition,
 author = {Christiansen, M.},
 journal = {Transportation science},
 number = {1},
 pages = {3--16},
 publisher = {INFORMS},
 title = {Decomposition of a combined inventory and time constrained ship routing problem},
 volume = {33},
 year = {1999}
}

@article{coelho2012inventory,
 author = {Coelho, L.C. and Cordeau, J. and Laporte, G.},
 journal = {Computers \& Operations Research},
 number = {11},
 pages = {2537--2548},
 publisher = {Elsevier},
 title = {The inventory-routing problem with transshipment},
 volume = {39},
 year = {2012}
}

@article{coelho2014survey,
 author = {Coelho, L. and Cordeau, J. and Laporte, G.},
 journal = {Transportation Science},
 number = {1},
 pages = {1--19},
 title = {{Thirty years of inventory routing}},
 volume = {48},
 year = {2014}
}

@inproceedings{diniz2020efficient,
 author = {Diniz, P. and Martinelli, R. and Poggi, M.},
 booktitle = {Combinatorial Optimization: 6th International Symposium, ISCO 2020, Montreal, QC, Canada, May 4--6, 2020, Revised Selected Papers 6},
 organization = {Springer},
 pages = {273--285},
 title = {An efficient matheuristic for the inventory routing problem},
 year = {2020}
}

@article{gaur2004periodic,
 author = {Gaur, V. and Fisher, M.L.},
 journal = {Operations Research},
 number = {6},
 pages = {813--822},
 publisher = {INFORMS},
 title = {A periodic inventory routing problem at a supermarket chain},
 volume = {52},
 year = {2004}
}

@article{greif2024combinatorial,
 author = {Greif, T. and Bouvier, L. and Flath, C.M. and Parmentier, A. and Rohmer, S.U. and Vidal, T.},
 journal = {arXiv preprint arXiv:2402.04463},
 title = {Combinatorial Optimization and Machine Learning for Dynamic Inventory Routing},
 year = {2024}
}

@article{guimaraes2019two,
 author = {Guimar{\~a}es, T.A. and Coelho, L.C. and Schenekemberg, C.M. and Scarpin, C.T.},
 journal = {Computers \& Operations Research},
 pages = {220--233},
 publisher = {Elsevier},
 title = {The two-echelon multi-depot inventory-routing problem},
 volume = {101},
 year = {2019}
}

@inproceedings{gunawan2019simulated,
 author = {Gunawan, A. and Vincent, F.Y. and Widjaja, A.T. and Vansteenwegen, P.},
 booktitle = {2019 IEEE 15th International Conference on Automation Science and Engineering (CASE)},
 organization = {IEEE},
 pages = {691--696},
 title = {Simulated annealing for the multi-vehicle cyclic inventory routing problem},
 year = {2019}
}

@article{hendel2022adaptive,
 author = {Hendel, G.},
 journal = {Mathematical Programming Computation},
 number = {2},
 pages = {185--221},
 publisher = {Springer},
 title = {Adaptive large neighborhood search for mixed integer programming},
 volume = {14},
 year = {2022}
}

@article{lin1973effective,
 author = {Lin, S. and Kernighan, B.},
 journal = {Operations Research},
 number = {2},
 pages = {498--516},
 publisher = {INFORMS},
 title = {An effective heuristic algorithm for the traveling salesman problem},
 volume = {21},
 year = {1973}
}

@article{mahjoob2022modified,
 author = {Mahjoob, M. and Fazeli, S.S. and Milanlouei, S. and Tavassoli, L.S. and Mirmozaffari, M.},
 journal = {Sustainable Operations and Computers},
 pages = {1--9},
 publisher = {Elsevier},
 title = {A modified adaptive genetic algorithm for multi-product multi-period inventory routing problem},
 volume = {3},
 year = {2022}
}

@article{ohmori2021multi,
 author = {Ohmori, S. and Yoshimoto, K.},
 journal = {Uncertain Supply Chain Management},
 number = {2},
 pages = {351--362},
 title = {Multi-product multi-vehicle inventory routing problem with vehicle compatibility and site dependency: A case study in the restaurant chain industry},
 volume = {9},
 year = {2021}
}

@article{coelho2020variable,
  title={A variable MIP neighborhood descent for the multi-attribute inventory routing problem},
  author={Coelho, L. C. and De Maio, A. and Laganà, D.},
  journal={Transportation Research Part E: Logistics and Transportation Review},
  volume={144},
  pages={102137},
  year={2020},
  publisher={Elsevier}
}

@article{feng2025bi,
  title={Bi-objective inventory routing problem with uncertain demand: a data-driven robust optimisation approach},
  author={Feng, Yuqiang and Che, Ada and Lei, Jieyu},
  journal={International Journal of Production Research},
  volume={63},
  number={11},
  pages={3885--3912},
  year={2025},
  publisher={Taylor \& Francis}
}

@article{larrain2017variable,
  title={A variable MIP neighborhood descent algorithm for managing inventory and distribution of cash in automated teller machines},
  author={Larrain, Homero and Coelho, Leandro C and Cataldo, Alejandro},
  journal={Computers \& Operations Research},
  volume={85},
  pages={22--31},
  year={2017},
  publisher={Elsevier}
}

@article{vincent2025three,
  title={A three-stage matheuristic for the blood stochastic inventory routing problem},
  author={Vincent, F Yu and Salsabila, Nabila Yuraisyah and Gunawan, Aldy and Siswanto, Nurhadi},
  journal={Transportation Research Part E: Logistics and Transportation Review},
  volume={200},
  pages={104143},
  year={2025},
  publisher={Elsevier}
}

@article{pacheco2023exponential,
 author = {Pacheco, T. and Martinelli, R. and Subramanian, A. and Toffolo, T.A. and Vidal, T.},
 journal = {Transportation Science},
 number = {2},
 pages = {463--481},
 publisher = {INFORMS},
 title = {Exponential-size neighborhoods for the pickup-and-delivery traveling salesman problem},
 volume = {57},
 year = {2023}
}

@article{prins2004simple,
 author = {Prins, C.},
 journal = {Computers \& Operations Research},
 number = {12},
 pages = {1985--2002},
 publisher = {Elsevier},
 title = {A simple and effective evolutionary algorithm for the vehicle routing problem},
 volume = {31},
 year = {2004}
}

@article{ropke2006adaptive,
 author = {Ropke, S. and Pisinger, D.},
 journal = {Transportation science},
 number = {4},
 pages = {455--472},
 publisher = {Informs},
 title = {An adaptive large neighborhood search heuristic for the pickup and delivery problem with time windows},
 volume = {40},
 year = {2006}
}

@article{rusdiansyah2005integrated,
 author = {Rusdiansyah, A. and Tsao, D.},
 journal = {Journal of Food Engineering},
 number = {3},
 pages = {421--434},
 publisher = {Elsevier},
 title = {An integrated model of the periodic delivery problems for vending-machine supply chains},
 volume = {70},
 year = {2005}
}

@article{skaalnes2023branch,
 author = {Sk{\aa}lnes, J. and Vadseth, S.T. and Andersson, H. and St{\aa}lhane, M.},
 journal = {Computers \& Operations Research},
 pages = {106353},
 publisher = {Elsevier},
 title = {A branch-and-cut embedded matheuristic for the inventory routing problem},
 volume = {159},
 year = {2023}
}

@article{skaalnes2024new,
 author = {Sk{\aa}lnes, J. and Ahmed, M.B. and Hvattum, L.M. and St{\aa}lhane, M.},
 journal = {European Journal of Operational Research},
 number = {3},
 pages = {992--1014},
 publisher = {Elsevier},
 title = {New benchmark instances for the inventory routing problem},
 volume = {313},
 year = {2024}
}

@article{Thompson1993,
author = {Thompson, P.M. and Psaraftis, H.N.},
journal = {Operations Research},
number = {5},
pages = {935--946},
title = {{Cyclic transfer algorithms for multi-vehicle routing and scheduling problems}},
volume = {41},
year = {1993}
}

@article{Capua2018,
author = {Capua, R. and Frota, Y. and Ochi, L.S. and Vidal, T.},
journal = {Journal of Heuristics},
number = {4},
pages = {667--695},
title = {{A study on exponential-size neighborhoods for the bin packing problem with conflicts}},
volume = {24},
year = {2018}
}

@incollection{Toth2008,
address = {New York},
author = {Toth, P. and Tramontani, A.},
booktitle = {The Vehicle Routing Problem: Latest Advances and New Challenges},
editor = {Golden, B.L. and Raghavan, S. and Wasil, E.A.},
pages = {275--295},
publisher = {Springer},
title = {{An integer linear programming local search for capacitated vehicle routing problems}},
year = {2008}
}

@article{Vidal2017b,
author = {Vidal, T.},
journal = {Operations Research},
number = {4},
pages = {992--1010},
title = {{Node, edge, arc routing and turn penalties: Multiple problems -- One neighborhood extension}},
volume = {65},
year = {2017}
}

@article{Balas2001,
author = {Balas, E. and Simonetti, N.},
journal = {INFORMS Journal on Computing},
number = {1},
pages = {56--75},
title = {{Linear time dynamic-programming algorithms for new classes of restricted TSPs: A computational study}},
volume = {13},
year = {2001}
}

@article{Toffolo2019,
author = {Toffolo, T.A.M. and Vidal, T. and Wauters, T.},
journal = {Computers \& Operations Research},
pages = {118--131},
title = {{Heuristics for vehicle routing problems: Sequence or set optimization?}},
volume = {105},
year = {2019}
}

@article{vidal2012hybrid,
 author = {Vidal, T. and Crainic, T.G. and Gendreau, M. and Lahrichi, N. and Rei, W.},
 journal = {Operations Research},
 number = {3},
 pages = {611--624},
 publisher = {INFORMS},
 title = {A hybrid genetic algorithm for multidepot and periodic vehicle routing problems},
 volume = {60},
 year = {2012}
}

@article{vidal2013hybrid,
 author = {Vidal, T. and Crainic, T.G. and Gendreau, M. and Prins, C.},
 journal = {Computers \& operations research},
 number = {1},
 pages = {475--489},
 publisher = {Elsevier},
 title = {A hybrid genetic algorithm with adaptive diversity management for a large class of vehicle routing problems with time-windows},
 volume = {40},
 year = {2013}
}

@article{vidal2014unified,
 author = {Vidal, T. and Crainic, T.G. and Gendreau, M. and Prins, C.},
 journal = {European Journal of Operational Research},
 number = {3},
 pages = {658--673},
 publisher = {Elsevier},
 title = {A unified solution framework for multi-attribute vehicle routing problems},
 volume = {234},
 year = {2014}
}

@article{vidal2020survey,
 author = {Vidal, T. and Laporte, G. and Matl, P.},
 journal = {European Journal of Operational Research},
 pages = {401--416},
 title = {{A concise guide to existing and emerging vehicle routing problem variants}},
 volume = {286},
 year = {2020}
}

@article{vidal2022hybrid,
 author = {Vidal, T.},
 journal = {Computers \& Operations Research},
 pages = {105643},
 publisher = {Elsevier},
 title = {{Hybrid genetic search for the CVRP: Open-source implementation and SWAP* neighborhood}},
 volume = {140},
 year = {2022}
}

@article{vincent2021multi,
 author = {Vincent, F.Y. and Widjaja, A.T. and Gunawan, A. and Vansteenwegen, P.},
 journal = {Computers \& Industrial Engineering},
 pages = {107320},
 publisher = {Elsevier},
 title = {The multi-vehicle cyclic inventory routing problem: Formulation and a metaheuristic approach},
 volume = {157},
 year = {2021}
}

@article{archetti2022comparison,
  title={Comparison of formulations for the inventory routing problem},
  author={Archetti, C. and Ljubić, I.},
  journal={European Journal of Operational Research},
  volume={303},
  number={3},
  pages={997--1008},
  year={2022},
  publisher={Elsevier}
}

@Inbook{Bauschke2017,
author="Bauschke, H.H.
and Combettes, P.L.",
title="Infimal Convolution",
bookTitle="Convex Analysis and Monotone Operator Theory in Hilbert Spaces",
year="2017",
publisher="Springer International Publishing",
address="Cham",
pages="203--217"
}

@article{Vidal2016,
author = {Vidal, T.},
journal = {Computers \& Operations Research},
pages = {40--47},
title = {{Technical note: Split algorithm in O(n) for the capacitated vehicle routing problem}},
volume = {69},
year = {2016}
}

\newpage
\begin{APPENDICES}
\APPENDICES

\section{\textcolor{black}{Considering Possible Backorders}}
\label{appendix:backlog}

\textcolor{black}{
The IRP formulation considered in this paper assumes lost sales: that is, unmet demand in a given period is not carried over and instead incurs an immediate penalty cost. This reflects many real-world applications, such as retail, vending machines, or seasonal products, where backlogging is either impractical or undesirable. 
Nevertheless, from an algorithmic perspective, supporting backorders requires only minimal modifications to the DS operator and no modification of HGS. In particular, this requires enlarging the DP state space and adapting the cost-update function.}

\textcolor{black}{
\noindent
\textbf{State Space.}
In the current model, the inventory level at the start of each period $I_i^t$ is restricted to the range $[0, U_i]$, where $U_i$ is the retailer's capacity. To allow backorders, this domain should be expanded to $[-B_{\max}, U_i]$, where $B_{\max}$ is the maximum backlog allowed for retailer $i$ (e.g., the sum of all demands). This change increases the number of DP states and thus raises the computational complexity of the operator from $O(U_iKH)$ to $O((U_i + B_{\max})KH)$ per retailer.}

\textcolor{black}{
\noindent
\textbf{Cost Function Modification.}
The rest of the DP structure, including the transition functions $f_1$~and~$f_2$, remains unchanged. The only modification occurs in the final cost function update, where the stockout penalty is replaced by a per-period backlog cost. Specifically, we define:
\begin{equation}
C_t(I_i^t) = \hat{C}_t(I_i^t) + p_i \cdot \max(0, -I_i^t),
\end{equation}
where $p_i$ is the backlog penalty per unit per day for retailer~$i$, and $\hat{C}_t$ is the continuation cost computed recursively. The term $\max(0, -I_i^t)$ captures the amount of backlog at the start of period $t$.}

\textcolor{black}{
\noindent
\textbf{Backtracking.}
In the backtracking phase, the delivery quantity $q_i^t$ must satisfy both the current demand and any carried-over backlog. If $I_i^{t-1} < 0$, the algorithm will naturally allocate enough delivery in period $t$ to reduce or clear the backlog. This behavior is seamlessly supported by the modified state space and cost function, requiring no further structural change.
This simple extension allows the DS operator to balance routing, inventory holding, and backlog costs, making it applicable to contexts where backorders are acceptable or even necessary.}

\end{APPENDICES}

\end{document}